\documentclass[journal]{IEEEtran}
\usepackage{changepage, graphicx}
\usepackage{algorithm, multirow}
\usepackage{algorithmic, CJK}
\usepackage{amsmath, indentfirst}
\usepackage{amssymb}
\usepackage{float}
\usepackage{subfigure}
\usepackage{cite}
\usepackage{url}
\usepackage{color, soul}
\usepackage{color}
\usepackage{framed}
\usepackage{array}
\usepackage{mathrsfs}
\usepackage{booktabs}
\usepackage[colorlinks,urlcolor=black,linkcolor=black,citecolor=black,anchorcolor=black,filecolor=black,menucolor=black,runcolor=black]{hyperref} 

\definecolor{shadecolor}{rgb}{1,0.94509804,0}
\soulregister\cite7 
\soulregister\citep7 
\soulregister\citet7 
\soulregister\ref7 
\soulregister\pageref7 
\soulregister\math7 
\soulregister\begin7 

\makeatletter

\begin{document}
\title{A Split-Window Transformer for Multi-Model Sequence Spammer Detection using Multi-Model Variational Autoencoder}
\author{
	Zhou~Yang, Yucai~Pang, Hongbo Yin, Yunpeng~Xiao 
	\thanks{This paper is partially supported by the National Natural Science Foundation of China (Grant No.62072066, 62006032), the Key Cooperation Project of Chongqing Municipal Education Commission(Grant No.HZ2021008) and Youth Innovation Group Support Program of ICE Discipline of CQUPT (Grant No.SCIE-QN-2022-05).\emph{(Corresponding author: Yucai Pang and Yunpeng Xiao.)}}
	
	\thanks{Z. Yang, Y. Pang, and Y. Xiao are with the School of Communications and Information Engineering, Chongqing University of Posts and Telecommunications, Chongqing, 400065, China (e-mail: yzhoul392@gmail.com;  pangyc@cqupt.edu.cn; xiaoyp@cqupt.edu.cn; ).}
	
	\thanks{H. Yin is with the School of Information and Communication Engineering, University of Electronic Science and Technology of China, Chengdu, 611731, China (e-mail: yinhub@yeah.net).}
}
\markboth{} %
{Shell \MakeLowercase{\textit{et al.}}: Bare Demo of IEEEtran.cls for IEEE Journals}
\maketitle
\begin{abstract}
	This paper introduces a new Transformer, called MS$^2$Dformer, that can be used as a generalized backbone for \textbf{m}ulti-modal \textbf{s}equence \textbf{s}pammer \textbf{d}etection. Spammer detection is a complex multi-modal task, thus the challenges of applying Transformer are two-fold. Firstly, complex multi-modal noisy information about users can interfere with feature mining. Secondly, the long sequence of users' historical behaviors also puts a huge GPU memory pressure on the attention computation. To solve these problems, we first design a user behavior Tokenization algorithm based on the multi-modal variational autoencoder (MVAE). Subsequently, a hierarchical split-window multi-head attention (SW/W-MHA) mechanism is proposed. The split-window strategy transforms the ultra-long sequences hierarchically into a combination of intra-window short-term and inter-window overall attention. Pre-trained on the public datasets, MS$^2$Dformer's performance far exceeds the previous state of the art. The experiments demonstrate MS$^2$Dformer's\footnote{\url{https://github.com/yzhouli/MSDformer.}} ability to act as a backbone.
	
\end{abstract}
\section{Introduction}
\IEEEPARstart{S}pammers are important promoters of directing social opinion. Users who create spam or fake news for a long time are defined as spammers. Information dissemination is usually carried out by graph structures. Therefore, the graph neural network (GNN) becomes a generalized backbone for identifying spammers. Subsequently, GNN has been further developed. By combining the ideas of Convolution (GCN\cite{li2019spam}), Attention (GAT\cite{zhang2023detecting, jiang2024learning} and Graph Transformer\cite{chen2024gnn}), and Sampling (Graph-SAGE\cite{zhang2024predicting}), the performance of GNN is further improved.
\begin{figure}[htbp]
	\center{\includegraphics[width=1\linewidth] {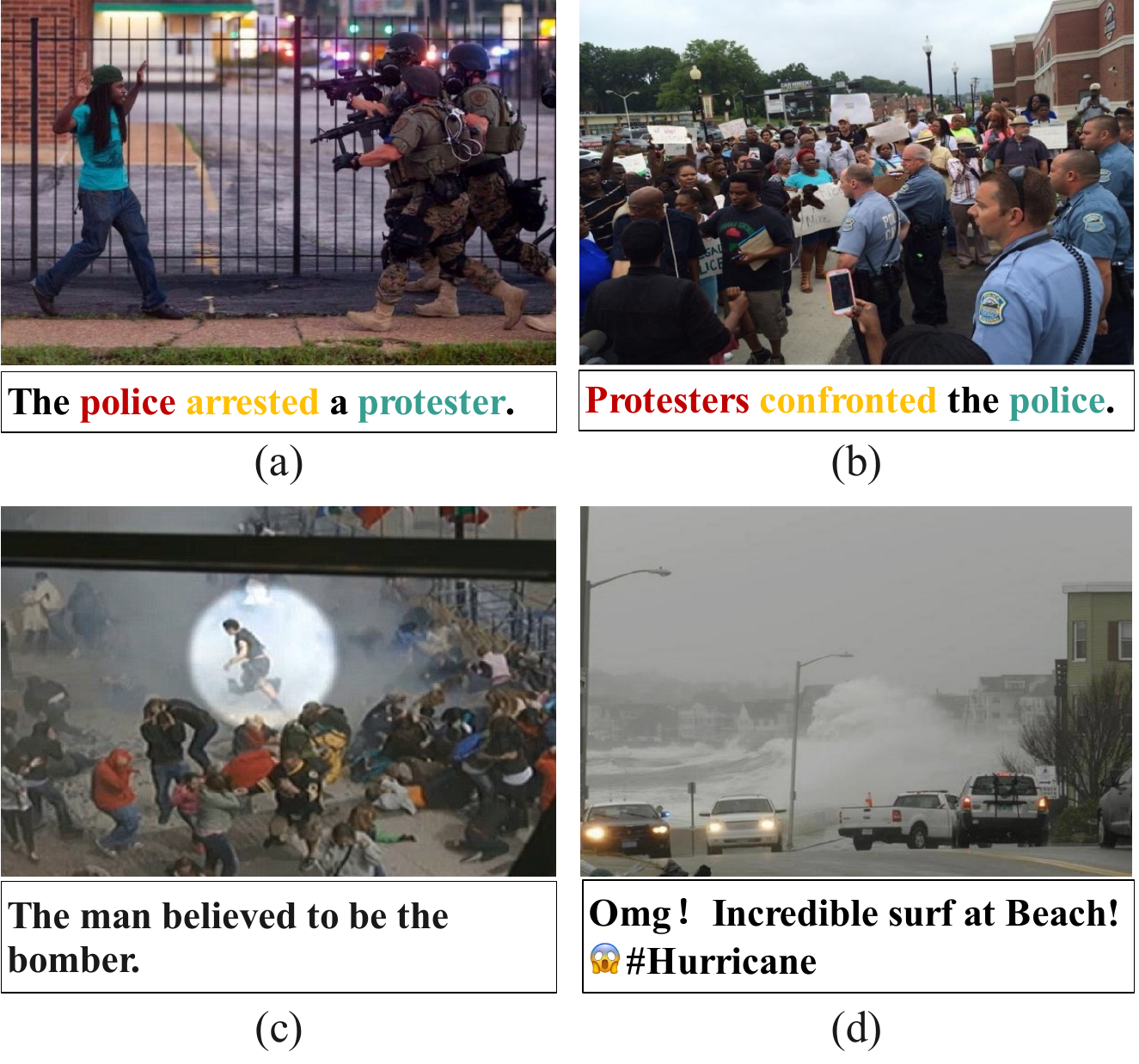}} 
	\caption{Some examples of complex cross-modal feature mining challenges. (a) and (b) are both standard cases, i.e., the core argument can be identified in the text modal, and the multi-modal provides auxiliary features. (c) the text modal does not directly provide the point of view, so further alignment and mining of the core argument in conjunction with the image is required. (d) the situation is most common in social behavior. User behavior from the text aspect is often accompanied by noise features, i.e., emojis and non-common characters (@, \#, and //, etc.). In special cases, it is also mixed with URL linking to elaborate the argument.}
	\label{fig-inspire}
\end{figure}
\begin{figure*}[h]
	\center{\includegraphics[width=1\linewidth]  {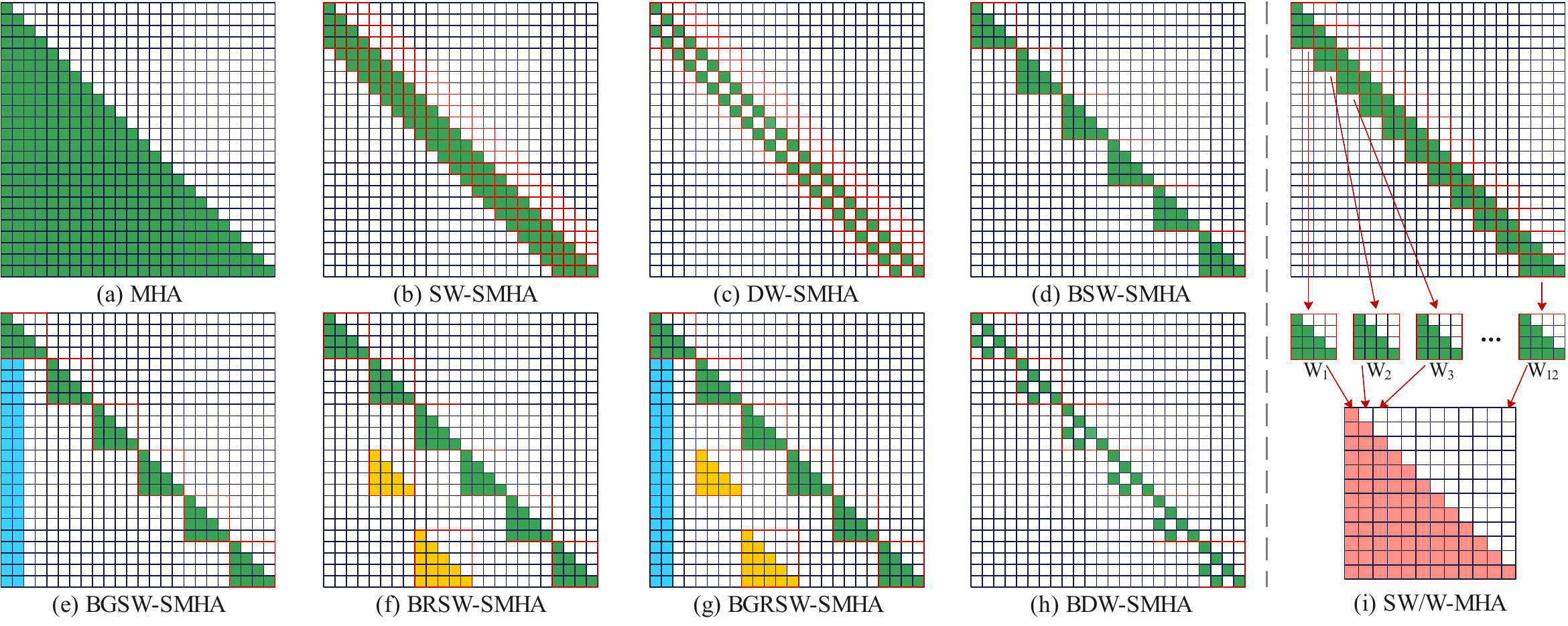}} 
	\caption{Inspiration for hierarchical attention mechanisms. (a) is the classical multi-head attention mechanism (MHA). (b-h) are sparse multi-head attention (SMHA) for solving ultra-long sequence modeling. (b) and (c) are SMHAs based on split windows. The core idea of them all is to limit the receptive field of an individual element, thus reducing the computational effort of $QK^\text{T}$. Among them, (c) expands the windowed receptive field similarly to the dilated convolution. Because the CPU dominates the windowing process, the full SMHA computation is slow (see Table \ref{table-ab-memory}). To solve this problem, the researchers adjusted the sliding distance to be consistent with the window length, thus proposing the block split-window mechanism (see (d) and (h)). Meanwhile, considering the importance of CLS tokens, Longformer (e\cite{beltagy2020longformer}) proposes global attention based on (d). Subsequently, random sampling is also added (see (f) and (g)).}
	\label{fig-MHAs}
\end{figure*}
\par On the other hand, scholars believe that spammers usually use sudden and short-term coherent behaviors to guide public opinion. Subsequently, they make quick disguises and gather followers with the help of behaviors such as spreading positive energy to prepare for the subsequent guidance. Meanwhile, they also use outdated information as a vehicle and add new claims to guide new public opinion. Moreover, the interval between outdated information and current activities is very long (see Fig. \ref{fig-ultra-long-term}). Therefore, sequence modeling strategies have also recently been applied to the current task. Unlike the GNN backbone, this strategy focuses on the hidden correlations between users' historical long-term and short-term behaviors. If a single historical behavior is considered a Token, the entire problem of modeling historical behavioral sequences can be transformed into the problem of natural language processing (NLP). In the field of NLP, Transformer is designed for sequence modeling and translation tasks, and its distinguishing feature is the use of attention to model long-term or short-term dependencies in the data. This is highly consistent with the sequence modeling needs of the task at hand.
\par However, there are differences between NLP and spammer detection tasks. Therefore, there are two issues that need to be addressed to apply the well-established Transformer backbone in the NLP domain for spammer detection tasks.

\begin{itemize}
	\setlength{\itemsep}{-1pt}
	\setlength{\parsep}{0pt}
	\setlength{\parskip}{0pt}
	\item[1)]
	The problem of multi-modal feature mining. Social networks are typically complex networks. Therefore, the representation carriers of user behaviors are also diverse, i.e., including text and multi-modal auxiliary features. In NLP tasks, researchers only need to consider how to mine semantic and contextual information. However, the problem of multi-modal feature mining and cross-modal alignment must also be considered in the spammer detection task (see Fig. \ref{fig-inspire}).
	\item[2)]
	The problem of representing ultra-long historical behavioral sequences. In the field of NLP, the classical MHA-based Transformer architecture supports token lengths typically between 512 and 8192 (BERT (512 tokens\cite{devlin2018bert}), GPT-2 (1024 tokens\cite{radford2019language}), GPT-3 (2048 tokens\cite{brown2020language}), GPT-4 (8192 tokens\cite{achiam2023gpt}), etc.). However, the length of historical user behavior sequences in the spammer detection task far exceeds this threshold. It has been experimentally verified (see Table 1) that the sequence length of 16, 384, or 36,768 is the most effective. At this point, the problem to be considered is solving the memory explosion caused by the $QK^T$ matrix during the multi-head attention computation. Meanwhile, miniaturized design must be considered because of the excessive number of daily social users. With the help of SMHA (see Fig. \ref{fig-MHAs}), the Transformer structure can handle ultra-long sequences. However, SMHA fails to meet the requirements at two levels of running efficiency and memory consumption (see Table \ref{table-ab-memory}).
\end{itemize}

\par To solve the above problems, we construct a novel Transformer, called MS$^2$Dformer, that can be used as a generalized backbone for \textbf{m}ulti-modal \textbf{s}equence \textbf{s}pammer \textbf{d}etection. Meanwhile, the variational autoencoder and hierarchical sparse attention mechanism are combined to solve complex multi-modal feature mining and ultra-long sequence modeling. Subsequently, the Transformer backbone is built to detect spammers. The contributions of this article are shown as follows:

\begin{itemize}
	\setlength{\itemsep}{-1pt}
	\setlength{\parsep}{0pt}
	\setlength{\parskip}{0pt}
	\item[1)]
	A Tokenization algorithm is proposed for user behavior based on a multi-modal variational autoencoder (MVAE). Firstly, pre-trained models (BERT\cite{wang2024utilizing} and ViT\cite{liu2021swin, han2022survey}) are utilized to embed multi-modal user behaviors. Secondly, dual-channel multi-modal feature encoding and cross-modal feature alignment are constructed. Subsequently, a dual-channel decoding component that shares cross-modal features is built\cite{wang2024revisiting, fang2023unsupervised}. Finally, the cross-modal encoded features are involved in feature reconstruction and sequence spammer detection.
	\item[2)]
	A hierarchical split-window multi-head attention (SW/W-MHA, see Fig. \ref{fig-MHAs} (i)) computational mechanism is proposed. Firstly, the split-window strategy is constructed to transform the ultra-long sequences into short-term attention within the window by hierarchical transformation. This action ensures the model can handle ultra-long Token sequences while further mining short-term spammers' bursty behaviors. Moreover, the windows overlap and slide to prevent omitting important short-term information (SW-MHA). Subsequently, the overall attention between windows is built to aggregate the long-term latent behaviors (W-MHA) deeply. SW/W-MHA significantly reduces the explicit explosion potential of MHA computation and accelerates the inference.
	\item[3)]
	Pre-training is performed in two publicly available datasets. MS$^2$Dformer significantly outperforms the previous state of the art, achieving an accuracy improvement of +6.9/5.2\%. In the consumer-grade platform (RTX 4060 (16GB)), the model is built with more than 53M parameters and runs efficiently (see Table \ref{table-ab-memory}). The experiments demonstrate MS$^2$Dformer's ability to act as a multi-modal spammer detection backbone.
\end{itemize}

\par The organization of the remaining sections of this article is shown as follows: Section II describes recent work related to the tasks of spammer detection, multi-modal variational autoencoder, and long sequence Transformer architecture. Section III describes the relevant definitions needed to construct the MS$^2$Dformer model. Section IV focuses on describing the details of the model. Section V compares the performance of the MS$^2$Dformer model with the current state-of-the-art algorithms pre-trained on a public dataset. Meanwhile, the ablation study is constructed to validate the rationality of the model components. Section VI summarizes the entire article and looks forward to future research work.

\section{Related Work}
\par Spammers guide public opinion by sending spam (also called fake news or rumors by some scholars). Therefore, the academic definition of spammer detection contains two sub-tasks: short-term burst and long-term hidden user detection. In the former, several short-term behaviors of a user are combined to identify whether the user is a spammer or not. This task is usually called spam (fake news or rumor) detection. In this case, the model is usually constructed by combining the spread space and the GNN backbone because the user has fewer short-term behaviors. The latter is more suitable for identifying long-term potential users. Meanwhile, it can also consider the task of sudden user detection. Moreover, with the arithmetic power improvement, multi-modal feature mining is combined in these two sub-tasks.
\par \textbf{GNN Backbone-Based Models:} Social diffusion processes are usually carried out in graphs. Therefore, GNN-based models are popularly used for the task at hand. For instance, Bian et al.\cite{Bian2020Rumor} used top-down and bottom-up bi-directional GCN models to solve the task of fake news recognition. Subsequently, Wei et al.\cite{wei2024modeling} introduced randomization theory based on Bian et al. Consequently, this strategy dramatically enhances the generalizability of the fake news recognition model. Meanwhile, the GNN backbone incorporating Markov fields\cite{deng2023markov} was also applied to the task of fake maker detection. It has been proven that the GNN-based model has reached an almost unsurpassable performance in terms of recognition accuracy. However, as the historical user behavior increases, i.e., when identifying long-hidden spammers, the GNN-based model consumes a large amount of GPU memory to derive suspicious behaviors. With hundreds of millions of social behaviors every day, models must be built with GPU memory in mind and ideally run on consumer GPUs. Therefore, the GNN backbone is not the most effective in latent spammer detection against the background of ultra-long historical behavioral sequences.
\par \textbf{Sequence-Based Models:} Social network spread space contains graph structure and temporal sequence information. Therefore, some researchers try constructing a detection model using a sequence modeling strategy. By combining temporal features, Yang et al.\cite{Yang2024Topic} constructed a spreading audio based on emotional entropy. Subsequently, fake news is recognized with the help of audio classification techniques. Similarly, Ma et al.\cite{ma2016detecting} proposed a temporal feature-based modeling strategy for propagated response information and identified the fake news with the help of RNN (Recurrent Neural Network). Subsequently, Ma et al.\cite{ma2021improving} constructed a GAN (Generative Adversarial Network) style model based on TD-RvNN\cite{ma2016detecting}. In this case, the Transformer model is used to model time-propagated sequences, and it is used as a generator. Subsequently, the RNN model is introduced as a discriminator to identify fake news. At this point, the authors mainly address modeling the short-term behavior-derived spread space. Therefore, they used the well-established Transformer architecture of the NLP domain that supports 512 tokens. Their model removes behaviors exceeding 512 tokens when confronted with ultra-long user sequences. Meanwhile, temporal spreading graphs were also applied. For instance, Sun et al.\cite{Sun2022ddgcn} mined the structural information of the propagation subspace based on GCN models. Subsequently, a bidirectional fusion strategy for structural features of the spread subspace was constructed based on temporal features. They did not address the training and deployment of temporal GNN backbone networks on consumer GPU platforms.
\par \textbf{Multi-Modal Models:} With the rapid development of computer and communication technologies, online social behaviors are no longer limited to text models. Therefore, multi-modal data mining and cross-modal alignment have become a hot topic in current research. In short-term burst detection, researchers have focused on the multi-modal domain. For instance, Wang et al.\cite{wang2023cross} introduced adversarial learning to align cross-modal features. Zhang et al.\cite{zhang2024reinforced} introduced a reinforcement learning strategy to learn cross-modal features. However, multi-modal features have not been widely used in the long-term hidden user detection sub-task. For instance, Qu et al.\cite{qu2024temporal} considered using a multi-modal modeling strategy. However, they only transformed the text into three channels of similar visual information, not accurate multi-modal data. To this end, we construct a generalized backbone for multi-modal spammer detection, called MS$^2$Dformer, from a sequence modeling perspective. Firstly, two-channel feature mining and alignment are constructed using multi-modal variational auto-encoding. Subsequently, sequence features are deeply quantified based on a hierarchical split-window attention mechanism.
\section{Problem Definition}
\subsection{Related Definitions}
\par Spammer detection is a typically multi-modal and complex task. Therefore, the input sequence data for the MS$^2$Dformer model is defined in this section starting from multi-modal feature extraction of text and images.
\par \textbf{Definition 1.} Text modal feature \begin{math} T^v \end{math}
\par As shown in Fig. \ref{fig-inspire}, the data from the text model in the user's historical behavior often provides very important data. Therefore, in this section, a pre-trained BERT\footnote{\url{https://huggingface.co/google-bert/bert-base-uncased.}} model with only an encoder structure is used to embed the text. The equation is shown as follows:
\begin{equation}
	T^v = \text{BERT}(\text{CLS}^{\text{T}}+T) \in \mathbb{R}^{768}
	\label{eq-1}
\end{equation}
where $T$ represents the text data. In the NLP domain, the textual data is concatenated with the “CLS” token at the beginning of the sentence. Thus, $\text{CLS}^{\text{T}}$ represents this token. $T^v\in \mathbb{R}^{768}$ represents the embedded representation of the “CLS” token encoded by the pre-trained BERT model.
\par \textbf{Definition 2.} Image modal feature \begin{math} I^v \end{math}
\par Guiding public opinion through textual modalities alone can cause the target group to trigger a crisis of confidence. Ultimately, it leads to a failure of provocation. Therefore, a spammer may provide multi-modal information, such as edited images, to testify his viewpoint. Thus, while considering text features, image features are equally indispensable. For this purpose, a pre-trained ViT\footnote{\url{https://huggingface.co/google/vit-base-patch16-224.}} model is used to extract image features. The equation is shown as follows:
\begin{equation}
	I^v = \text{ViT}(I) \in \mathbb{R}^{768}
	\label{eq-2}
\end{equation}
where $I$ represents the input image data. $I^v\in \mathbb{R}^{768}$ represents the embedded representation of the “CLS” token encoded by the pre-trained ViT model. Because ViT adds “CLS” tokens inside the model, they are not represented in Eq. (\ref{eq-2}).
\par \textbf{Definition 3.} User Historical Behavior Sequence \begin{math}S=\{(T^v_0, I^v_0),..., (T^v_l, I^v_l)\}\end{math}
\par This paper proposes a serialized model for spammer detection. Therefore, it is first necessary to transform a sequence of historical user behaviors $U=\{b_0, ..., b_l\}$ into a standard representation that the model can process. As shown in Fig. \ref{fig-inspire}, any behavior in the sequence contains both text and image modal data, i.e., $b_i=(T_i, I_i)$, and $i \in [0, l]$. In particular, a history behavior $b_i$ exists that fills an empty image when it does not contain image data. Subsequently, the sequence of history behaviors $\{b_0, ..., b_l\}$ into standard form. The equation is shown as follows:	
\begin{equation}
	\begin{split}
	S&=\text{BERT}(\{T_0, ..., T_l\})\ \text{and}\ \text{ViT}(\{I_0, ..., I_l\})\\
	   &=\{(T^v_0, I^v_0),..., (T^v_l, I^v_l)\} \in \mathbb{R}^{l\times2\times768}
	\label{eq-3}
	\end{split}
\end{equation}
where $S \in \mathbb{R}^{l\times2\times768}$ represents the standard sequence of inputs for user $U$. $l$ represents the length of the sequence supported by the model. The user's real behavior sequence will be truncated when its length exceeds $l$. On the other hand, insufficient sequences supplement the empty behaviors. The effect of these empty behaviors is eliminated in the model by the MASK mechanism. Meanwhile, to accelerate the extraction of multi-modal features, $\text{BERT}(\{T_0, ... , T_l\})$ and $\text{ViT}(\{I_0, ... , I_l\})$ strategies are employed to parallelize the extraction of features.
\subsection{Problem Formulation}
\par To solve the problem of serialized multi-modal spammer detection mathematically, we propose a new Transformer, called MS$^2$Dformer. The model models this complex problem in four separate stages. Thus, the overall representation of the model is as follows:
\begin{eqnarray}\left. {\begin{array}{*{20}{l}}
			U=\{b_0, ... , b_l\}\\
			b_i=(T_i, I_i)\\
	\end{array}} \right\} \Rightarrow \text{MS}^2\text{Dformer} \Rightarrow {P\{s,n \mid user\}}
\end{eqnarray}
\begin{figure*}[h]
	\center{\includegraphics[width=1\linewidth]  {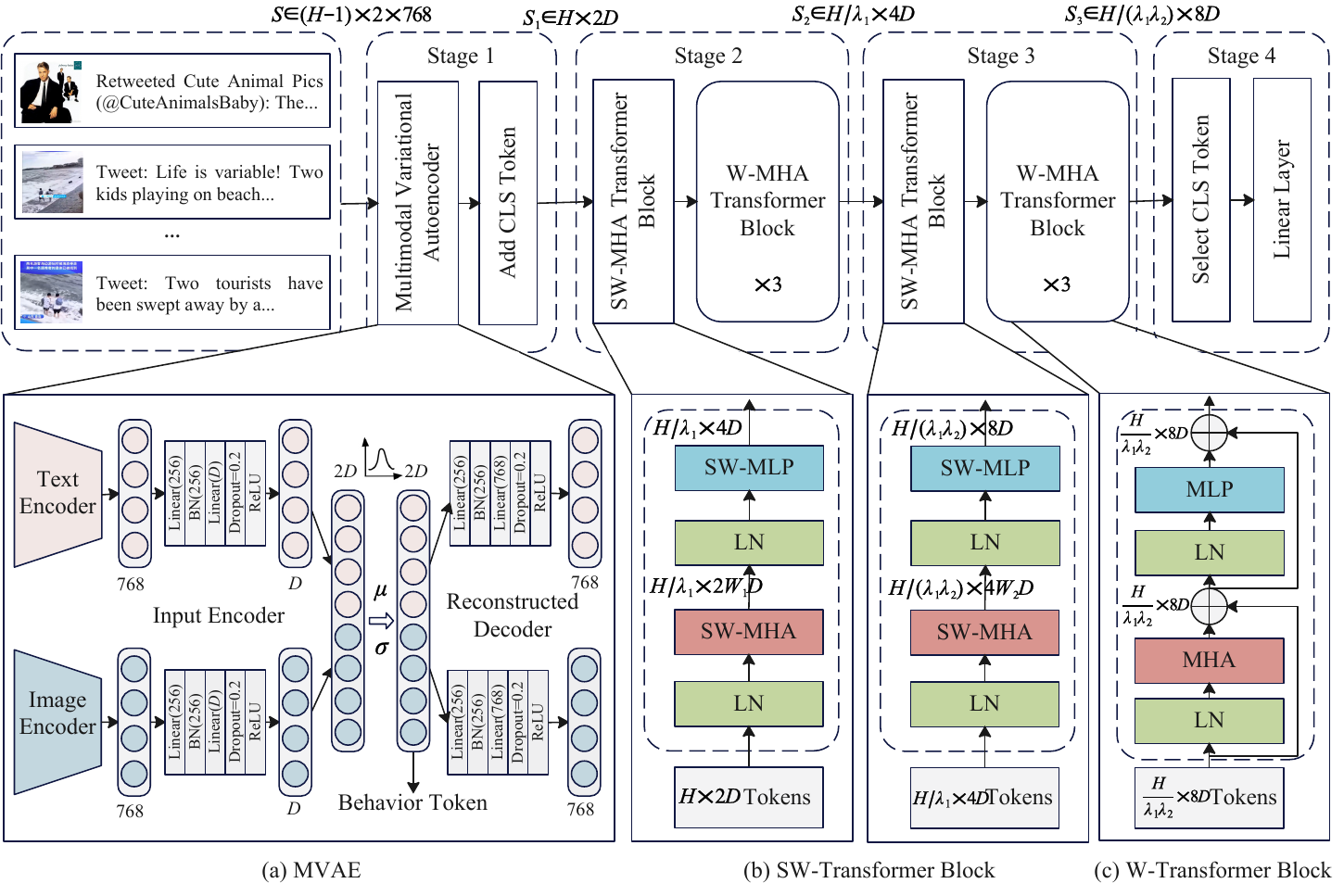}} 
	\caption{The framework of the MS$^2$Dformer\_B model (MS$^2$Dformer Base Version, see Eq. (\ref{eq-base})).}
	\label{fig-2}
\end{figure*}
\subsubsection{Model Input}\
\par The input data for the MS$^2$Dformer model is shown as follows:
\begin{itemize}
	\setlength{\itemsep}{-1pt}
	\setlength{\parsep}{0pt}
	\setlength{\parskip}{0pt}
	\item
	The sequence of historical user behaviors $U=\{b_0, ... , b_l\}$. $l$ represents the length of the sequence supported by the model.
	\item
	Individual user behavior $b_i=(T_i, I_i)$. The $i$-th behavior of user $U$ contains text $T_i$ and image $I_i$, and $i \in [0, l]$.
\end{itemize}

\subsubsection{Model Output}\ 
\par Based on the model input data from the previous section, the MS$^2$Dformer model (see Fig. \ref{fig-2}) needs to solve the following problems in stages.
\begin{itemize}
	\setlength{\itemsep}{-1pt}
	\setlength{\parsep}{0pt}
	\setlength{\parskip}{0pt}
	\item
	\textbf{Stage 1:} Multi-modal hidden features $S_1 \in \mathbb{R}^{H\times2D}$. Firstly, the model input sequence $U=\{b_0, ... , b_l\}$ after Eq. (\ref{eq-3}) to the embedding representation $S=\{(T^v_0, I^v_0),... , (T^v_l, I^v_l)\} \in \mathbb{R}^{l\times2\times768}$. Secondly, a two-channel multi-modal variational autoencoder (MVAE) is constructed from the variational autoencoder (VAE). After MVAE, $S \in \mathbb{R}^{l\times2\times768}$ is transformed into $\mathbb{R}^{l\times2D}$. $D$ represents the dimension of the MVAE embedding part. Finally, the classification token CLS$^{\text{S}}$ is added. Thus, $S \in \mathbb{R}^{l\times2\times768}$ is transformed into $S_1 \in \mathbb{R}^{H\times2D}$, and $H=l+1$.
	\item		
	\textbf{Stage 2:} Solve the problem of modeling ultra-long sequences. Firstly, the hierarchical split-window attention mechanism (SW/W-MHA) is constructed. Subsequently, the SW-MHA Transformer block is constructed based on the intra-window attention mechanism. In this case, $S_1 \in \mathbb{R}^{H\times2D}$ is transformed by SW-MHA into $\mathbb{R}^{\frac{H}{\lambda_{1}}\times2W_{1}D}$. $\lambda_{1}$ represents the stride size of the window. $W_{1}$ represents the length of the window. Afterwards, the original multi-layer linear perceptron (MLP) is modified to SW-MLP. In addition, $S_1 \in \mathbb{R}^{\frac{H}{\lambda_{1}}\times2W_{1}D}$ is transformed by SW-MLP to $\in \mathbb{R}^{\frac{H}{\lambda_{1}}\times4D}$. Secondly, multiple W-MHA Transformer blocks are constructed based on the inter-window attention mechanism. In this case, the W-MHA Transformer block is used to deeply mine the inter-window sequence features. Therefore, no changes are made to the input data dimensions, i.e., $S_2 \in \mathbb{R}^{\frac{H}{\lambda_{1}}\times4D}$.
	\item	
	\textbf{Stage 3:} Solve the problem of deep sequence feature mining. Compared with Stage 2, there are two differences in Stage 3. Firstly, the window sliding step $\lambda_{2}$ set by the SW-MHA component in Stage 3 should be much smaller than $\lambda_{1}$. Secondly, the W-MHA component uses a deeper Block structure to mine sequence features. Overall, $S_2 \in \mathbb{R}^{\frac{H}{\lambda_{1}}\times4D}$ is transformed into $\mathbb{R}^{\frac{H}{\lambda_{1}\lambda_{2}}\times8D}$, i.e., $S_3 \in \mathbb{R}^{\frac{H}{\lambda_{1}\lambda_{2}}\times8D}$.
	\item		
	\textbf{Stage 4:} Solve the problem of spammer detection. Firstly, the classification token CLS$^{\text{S}} \in \mathbb{R}^{8D}$ in $S_3 \in \mathbb{R}^{\frac{H}{\lambda_{1}\lambda_{2}}\times8D}$ is selected. Lastly, the classification token is mapped to CLS$^{\text{S}} \in \mathbb{R}^{2}$ after two linear layers. Thus, the spammers are recognized by softmax function.
\end{itemize}

\section{MS$^2$Dformer Model}

\subsection{Overview}
\par A novel Transformer, called MS$^2$Dformer, is constructed that can be used as a generalized backbone for multi-modal sequence spammer detection. The model is divided into four stages to identify spammers (see Fig. \ref{fig-2}). Firstly, stage 1 completes tokenizing the user's historical behavior based on two-channel MVAE. Secondly, ultra-long sequences are modeled with stages 2 and 3, and sequence features are deeply mined. Finally, stage 4 employs a linear layer to mine classification token CLS$^{\text{S}}$ features and subsequently identifies spammers.
\begin{figure*}[h]
	\center{\includegraphics[width=1\linewidth]  {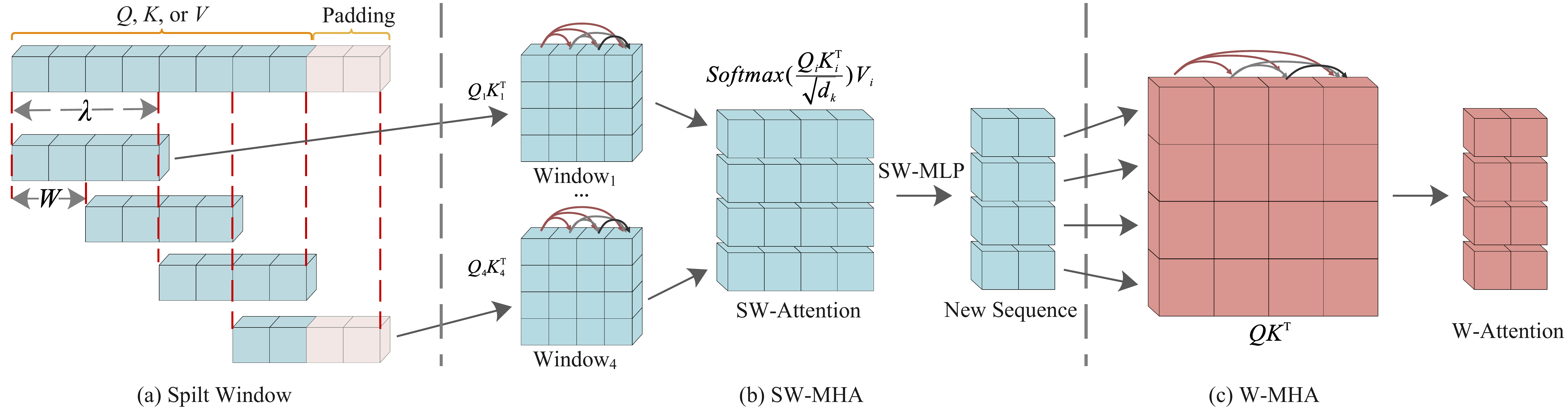}} 
	\caption{The case of SW/W-MHA.}
	\label{fig-SW-MHA}
\end{figure*}
\subsection{MVAE-based Historical Behavior Tokenization}
\par The primary problem to be solved for the serialized spammer identification model is how to tokenize the user's multi-modal historical behavior. To this end, we construct a two-channel MVAE-based tokenization strategy for user history behaviors. Subsequently, the serialization of the entire user history behavior is completed by the joint CLS$^{\text{S}}$ token. Therefore, this section contains three steps: \textit{Input Embedding}, \textit{MVAE}, and \textit{Behavior Tokenization}.
\par \textbf{Input Embedding:} Input embeddings generated by large pre-trained models are capable of encoding rich contextual information. To this end, we use BERT (see Eq. \ref{eq-1}) to obtain a single behavioral text feature embedding $T^v\in \mathbb{R}^{768}$ , and use ViT (see Eq. \ref{eq-2}) to obtain the image embedding $I^v\in \mathbb{R}^{768}$. Subsequently, the input embedding sequence $S \in \mathbb{R}^{l\times2\times768}$ of the MVAE component is constructed based on the original sequence $U$ (see Eq. \ref{eq-3}).
\par \textbf{MVAE:} Because the embedding already contains rich semantic information, there is no need to go through complex and deep layers to further extract contextual information in the MVAE's encoder. Therefore, we use two linear layers followed by Batch Normalization (BN) and Dropout (rate = 0.2) as the encoder structure for MVAE. BN is used to reduce the internal covariance bias by projecting the inputs to have mean zero and variance one. Subsequently, the formal representation of the encoder is shown below:
\begin{equation}
	z = \mu + \sigma \odot \epsilon
	\label{eq-4}
\end{equation}
\begin{equation}
	\mathcal{L} = \mathbb{E}_{q_\phi(z|x)}\left[\log p_\theta(x|z)\right] - \mathrm{D}_{\text{KL}}(q_\phi(z|x) \parallel p(z))
	\label{eq-5}
\end{equation}
where $\epsilon$ represents the noise sampled from the standard normal distribution $\mathcal{N}(0, I)$. $\odot$ represents the elemental product. $z$ denotes the latent variable. It is satisfying a Gaussian distribution. $\mu$ and $\sigma$ are the mean and variance of $z$, respectively. These parameters describe the conditional distribution $p_\theta(x|z)$ of the latent variable $z$. $\phi$ represents the parameters of the encoder. The objective of the decoder is to approximate the posterior distribution $q_\phi(z|x)$ to the posterior distribution $p_\theta(x|z)$.
\par In MVAE, the input data $x$ contains text embedding $T^v$ and image embedding $I^v$, i.e., $x=(T^v, I^v)$. Subsequently, the potential variables $z^\text{T}$ and $z^\text{I}$ are obtained after the encoding component of the two channels. Then, the composite feature $z=\text{concat}(z^\text{T}, z^\text{I})$ is obtained by combining the two-channel multi-modal potential features.
\par Afterwards, MVAE constructs two channel decoders for multi-modal feature reconstruction. The two encoders share a latent variable $z$. Therefore, the reconstruction loss function for MVAE is shown as follows:
\begin{equation}
	\begin{split}
	\mathcal{L}^\text{T} = \mathbb{E}_{q_\phi(z|T^v)}\left[\log p_\theta(T^v|z)\right] -\mathrm{D}_{\text{KL}}(q_\phi(z|T^v) \parallel p(z))
	\end{split}
	\label{eq-7}
\end{equation}
\begin{equation}
	\mathcal{L}^\text{I} = \mathbb{E}_{q_\phi(z|I^v)}\left[\log p_\theta(I^v|z)\right] - \mathrm{D}_{\text{KL}}(q_\phi(z|I^v) \parallel p(z))
	\label{eq-8}
\end{equation}
where $\mathcal{L}^\text{T}$ and $\mathcal{L}^\text{I}$ represent the loss for the text and image channels. $\mathbb{E}_{q_\phi(z|T^v)}\left[\log p_\theta(T^v|z)\right]$ and $\mathbb{E}_{q_\phi(z|I^v)}\left[\log p_\theta(I^v|z)\right]$ represent the reconstruction loss for the two channel. $\mathrm{D}_{\text{KL}}(q_\phi(z|T^v) \parallel p(z))$ and $\mathrm{D}_{\text{KL}}(q_\phi(z|I^v) \parallel p(z))$ represent the KL divergence Loss. $p(z)$ represents the standard normal distribution, i.e., $p(z)=\mathcal{N}(0, I)$.
\par \textbf{Behavior Tokenization:}
In MVAE, the latent variable $z$ is included across multi-modal features. Meanwhile, the complexities appearing in Fig. \ref{fig-inspire} are simulated by introducing the noise $\epsilon$. Subsequently, the interference features are eliminated by cyclically reconstructing the two-channel features. Therefore, the individual behavior corresponding to the latent variable $z$ is used as a token for the input sequence of stage 2.
\par Subsequently, a parallelization strategy is used to simultaneously extract the latent variables $z \in \mathbb{R}^{l\times2D}$ from the entire input sequence $S \in \mathbb{R}^{l\times2\times768}$. Where $D$ is a hyper-parameter indicating the final representation dimension of the MVAE encoder component. Later, the importance of classification tokens in the traditional Transformer architecture is taken into account. Therefore, we construct an all-zero matrix CLS$^{\text{S}} \in \mathbb{R}^{1\times2D}$ to act as the classification token. Lastly, we combine CLS$^{\text{S}}$ and $z$ to serve as the input sequence for stage 2, $S_1 \in \mathbb{R}^{H\times2D }$. In this case, to facilitate the computation, we define $H=l+1$.
\subsection{Ultra-long Behavior Sequence Mining}
\par \textbf{MHA:} The multi-head attention mechanism (MHA) has shown to be extremely powerful in sequence modeling tasks. Therefore, the MHA-based Transformer architecture is used as the basic backbone for sequence feature mining. Among them, the computational process of classical MHA is shown as follows:
\begin{equation}
	Q=XW^\text{Q},\ K=XW^\text{K},\ V=XW^\text{V}
	\label{eq-9}
\end{equation}
\begin{equation}
	\text{Attention}(Q, K, V)=\text{softmax}(\frac{QK^\text{T}}{\sqrt{d}})V
	\label{eq-10}
\end{equation}
\begin{equation}
	head_i=\text{Attention}(QW^\text{Q}_{i}, KW^\text{K}_{i}, VW^\text{V}_{i})
	\label{eq-11}
\end{equation}
\begin{equation}
	\begin{split}
		\widehat{X}&=\text{MHA}(Q, K, V)\\
		&=\text{Concat}(head_1,...,head_n)W^\text{M}
		\label{eq-12}
	\end{split}
\end{equation}
where $X$ represents the input sequence. $\widehat{X}$ represents the sequence after MHA computation. $D$ represents the final representation dimension of the MVAE encoder component. $H$ represents the sequence length. $Q$, $K$, and $V$ represent the query, key, and value of the sequence $X$, respectively. $\text{Attention}(Q, K, V)$ represents the attention computation for a individual head. $K^\text{T}$ represents the transpose of the $K$ matrix. $d$ represents the dimension of the key. $head_i$ represents the attention of the $i$-th head. $n$ represents the number of heads in the MHA. $W^\text{Q}_{i}$, $W^\text{K}_{i}$, $W^\text{V}_{i}$, and $W^\text{M}$ represent the trainable parameter matrices.
\par In the spammer detection task, the sequence of historical user behaviors is an ultra-long sequence. Therefore, $QK^\text{T}$ in traditional MHA computation (see Eq. 8) will construct an oversized matrix thus causing GPU memory crash (see Table \ref{table-ab-memory}). For this reason, we propose hierarchical split window attention. The core idea is to construct sliding windows $W$ in $QK^\text{T}$. Subsequently, the window $W$ is sliding sampled in steps $\lambda$. Subsequently, the length $H$ of the ultra-long sequence is transformed into $\frac{H}{\lambda}$. Thus, the problem of modeling ultra-long sequences is transformed into the problem of modeling standard sequences. In this case, the intra-window and inter-window attention is computed in chunks for $QK^\text{T}$ at two levels, thus approximating the attention of the entire sequence. To this end, a two-level windowed attention computation is proposed: \textit{SW-MHA} and \textit{W-MHA}.
\par \textbf{SW-MHA:} To avoid $QK^\text{T}$ oversized matrices, sliding window splitting is performed in the $Q$ and $K$ vectors. Subsequently, the intra-window attention of the window sequence is computed in parallel. The equation is shown as follows:
\begin{equation}
	\begin{split}
	\widehat{Q}=\text{SW}(XW^\text{Q}), \widehat{K}=\text{SW}(XW^\text{K}), \widehat{V}=\text{SW}(XW^\text{V})
	\end{split}
	\label{eq-13}
\end{equation}
\begin{equation}
	\text{SW-Att}_i=\text{softmax}(\frac{\widehat{Q}_{i}\widehat{K}_{i}^\text{T}}{\sqrt{d}})\widehat{V}_{i}, i\in [1, k]
	\label{eq-13-1}
\end{equation}
\begin{equation}
	\text{SW-Attention}(Q, K, V)=\text{concat}(\text{SW-Att}_1, ..., \text{SW-Att}_k)
	\label{eq-13-2}
\end{equation}
where $\text{SW}$ represents the sliding window splitting function, and the implementation process is shown in Fig. \ref{fig-SW-MHA} (a). $k$ represents the length of the sequence after the split window. It is assumed that $Q$, $K$ and $V$ are matrices of sequence length $H$ and embedding dimension $\eta$. Then, $Q$, $K$ and $V$ are all transformed into $\mathbb{R}^{\frac{H}{\lambda}\times W\times\eta}$ by the SW algorithm. $W$ and $\lambda$ represent the window length and sliding step, respectively.
\par Subsequently, the multi-head attention is computed in combination with Eq. (12). Finally, to satisfy the Transformer architecture calculation, $\widehat{X}$ is expanded in the last two dimensions. The equation is shown as follows:
\begin{equation}
	X^{\text{SW}}=\text{Flatten}(\widehat{X})
	\label{eq-14}
\end{equation}
where $X^{\text{SW}} \in \mathbb{R}^{\frac{H}{\lambda}\times W\eta}$ represents the feature matrix after intra-window attention (SW-MHA).
\par \textbf{W-MHA:} The computation process for inter-window attention uses traditional MHA (see Eqs. (\ref{eq-9}-\ref{eq-12})). Unlike MHA, the input sequence for inter-window attention is the features obtained after the calculation of intra-window attention. Therefore, inter-window attention does not change the shape of the input features.
\par \textbf{Transformer Block:} Subsequently, a comprehensive approximate representation of the overall attention is made based on the intra-window attention (SW-MHA) and inter-window attention (W-MHA). Given the effectiveness of the Transformer architecture, SW-Transformer Block and W-Transformer Block are constructed based on SW-MHA and W-MHA. The specific equations are shown as follows:
\begin{equation}
	\begin{split}
		X^{\text{SW}}&=\text{SW-MHA}(\text{LN}(X)))\\
		\xi&=\text{SW-MLP}(\text{LN}(X^{\text{SW}})))
	\end{split}
	\label{eq-15}
\end{equation}
\begin{equation}
	\begin{split}
		\widehat{\xi}&=\xi+\text{W-MHA}(\text{LN}(\xi)))\\
		X^{\text{O}}&=\widehat{\xi}+\text{MLP}(\text{LN}(\widehat{\xi})))
	\end{split}
	\label{eq-16}
\end{equation}
where Eq. (\ref{eq-15}) describes the architecture of SW-Transformer Block and Eq. (\ref{eq-16}) describes the architecture of W-Transformer Block. LN represents Layer Normalization. MLP represents standard multilayer linear perceptron. SW-MLP represents the multilayer linear perceptron specifically for SW-Transformer Block. Compared to MLP, SW-MLP acts not to increase the hidden state space but to decrease the feature dimension. For instance, the output feature $X^{\text{SW}} \in \mathbb{R}^{\frac{H}{\lambda}\times W\eta}$ of SW-MHA has an embedding dimension of $W\eta$. MLP does it by raising the dimension first to $2W\eta$ and subsequently lowering it to $W\eta$. The very large dimension space of $2W\eta$ may also cause GPU memory crash. Therefore, SW-MLP reduces the dimensionality first to $4\eta$ and subsequently to $2\eta$. In this case, SW-MLP prevents GPU memory explosion and also constructs the hidden state space. $\widehat{\xi}$ and $\xi$ represent intermediate variables. $X^{\text{O}}$ represents the Block output. Let's assume that the dimension of the input feature $X$ is $\mathbb{R}^{H\times\eta}$, then the dimensions of $\xi$, $\widehat{\xi}$, and $X^{\text{O}}$ are $\mathbb{R}^{\frac{H}{\lambda}\times2\eta}$.
\par \textbf{Stage 2 and 3:} The input sequence $X$ for Stage 2 is $S_1 \in \mathbb{R}^{H\times2D}$. Therefore, the main task of Stage 2 is to address the threat of GPU memory crashes caused by ultra-long sequences. To this end, a larger sliding step $\lambda_1$ of the window $W_1$ for intra-window feature mining (SW-Transformer Block) is required in Stage 2. Meanwhile, to mine inter-window relationship features, 3 layers of W-Transformer Block are set. Comparing with Stage 2, Stage 3 focuses on mining sequence features. Therefore, the sliding step $\lambda_2$ of the SW-Transformer Block window $W_2$ in Stage 2 does not need to be too large. Meanwhile, more W-Transformer Block needs to be set to mine sequence features.

\subsection{Spammer Detection}
\par Stage 1-3 are described in the previous two sections, respectively. In this section, the objective of Stage 4 is to identify spammers with an input sequence $S_3$ of dimension $\mathbb{R}^{\frac{H}{\lambda_1\lambda_2}\times8D}$. Firstly, the classification token $\text{CLS}^{\text{S}} \in 8D$ is selected from the sequence $S_3$. Subsequently, $\text{CLS}^{\text{S}}$ is input into two linear layers for feature learning and dimension transformation. Finally, spammers are identified with the help of softmax function. Thus, the final objective function of the model is shown as follows:
\begin{equation}
	\widehat{Y}=\text{Liner}(\text{Dropout}(\text{Liner}(\text{CLS}^{\text{S}})))
	\label{eq-17}
\end{equation}
where $\widehat{Y}$ represents the model output. Liner represents the linear layer. Dropout represents the Dropout layer, which is intended to adequately train the model parameters. Subsequently, the model classification loss uses the cross-entropy function. The equation is shown as follows:
\begin{equation}
	\mathcal{L}=-\frac{1}{N}\sum^{N}_{k}Y_k\text{log}(\widehat{Y_k})
	\label{eq-18}
\end{equation}
where $N$ represents the number of input samples in a batch. $Y_k$ represents the true labels of the samples. $\widehat{Y_k}$ represents the model prediction. Subsequently, the model total loss $\mathcal{L}^\text{Total}$ is obtained by combining the MVAE two-channel loss $\mathcal{L}^\text{I}$ and $\mathcal{L}^\text{T}$. The equation is shown as follows:
\begin{equation}
	\mathcal{L}^\text{Total}=\psi_1\times\mathcal{L}+\psi_2\times\mathcal{L}^\text{I}+\psi_3\times\mathcal{L}^\text{T}
	\label{eq-19}
\end{equation}
where $\psi_1$, $\psi_2$, and $\psi_3$ represent the decay factors of the three losses, respectively.

\subsection{Learning Algorithm}
\par A new Transformer, called MS$^2$Dformer, is proposed. Firstly, a two-channel MVAE for multi-modal behavior quantification is built based on the classical VAE (see Algorithm \ref{alg-1}). Subsequently, SW-Transformer Block based on the split-window attention mechanism is proposed (see Algorithm \ref{alg-2}).

\begin{algorithm}
	\renewcommand{\algorithmicrequire}{\textbf{Input:}}
	\renewcommand{\algorithmicensure}{\textbf{Output:}}
	\caption{{User Historical Behavior Tokenization}}
	\label{alg-1}
	\begin{algorithmic}[1]
		\REQUIRE \
		\par The sequence of historical user behaviors $U=\{b_0, ... , b_l\}$;
		\par Individual user behavior $b_i=(T_i, I_i)$;
		\ENSURE \ 
		\par Behavior tokens: $S_1$, Dual-channel loss: $\mathcal{L}^\text{T}$, and $\mathcal{L}^\text{I}$;
		
		\STATE $U \to \{(T^v_0, I^v_l),..., (T^v_l, I^v_l)\}=(T^v, I^v)$ by Eqs. (\ref{eq-1}-\ref{eq-3});
		\STATE $T^v \to z^\text{T}$ and $I^v \to z^\text{I}$ by dual-channel encoder;
		\STATE $z=\mu + \sigma \odot \epsilon=\text{concat}(z^\text{T}, z^\text{I})$;
		\STATE $z \to T^v_d$ and $z \to I^v_d$ by dual-channel decoder;
		\STATE $(T^v_d \to T^v) \to \mathcal{L}^\text{T}$ and $(I^v_d \to I^v) \to \mathcal{L}^\text{I}$;
		\STATE $S_1=\text{concat}(\text{CLS}^{\text{S}}, z)$;
		\STATE \textbf{return} $S_1$, $\mathcal{L}^\text{T}$, and $\mathcal{L}^\text{I}$;
	\end{algorithmic}
\end{algorithm}
\begin{algorithm}
	\renewcommand{\algorithmicrequire}{\textbf{Input:}}
	\renewcommand{\algorithmicensure}{\textbf{Output:}}
	\caption{{SW-MHA}}
	\label{alg-2}
	\begin{algorithmic}[1]
		\REQUIRE \
		\par Input Feature Sequence $S^\text{I} \in \mathbb{R}^{H\times\eta}$;
		\ENSURE \ 
		\par Output Feature Sequence $S^\text{O} \in \mathbb{R}^{\frac{H}{\lambda}\times W\eta}$;
		
		\STATE $Q=S^\text{I}W^\text{Q},\ K=S^\text{I}W^\text{K},\ V=S^\text{I}W^\text{V}$;
		\FOR{$i$ from $0$ to $H$ step $\lambda$)}
			\STATE $\widehat{Q}_i=Q[i\ \text{to}\ W], \widehat{K}_i=K[i\ \text{to}\ W], \widehat{V}_i=V[i\ \text{to}\ W]$;
		\ENDFOR \\
		// $\widehat{Q} \in \mathbb{R}^{\frac{H}{\lambda}\times W\times\eta}, \widehat{K} \in \mathbb{R}^{\frac{H}{\lambda}\times W\times\eta}, \widehat{V} \in \mathbb{R}^{\frac{H}{\lambda}\times W\times\eta}$  by window $W$ and sliding step $\lambda$;\\
		// To prevent cases where $H$ is not a multiple of $\lambda$, link $\lambda$ empty elements at the end of the sequence $S^\text{I}$, similar to the padding process used to CNNs, i.e. $S^\text{I} \in \mathbb{R}^{(H+\lambda)\times\eta}$;
		\STATE $head_i=\text{Attention}(\widehat{Q}W^\text{Q}_{i}, \widehat{K}W^\text{K}_{i}, \widehat{V}W^\text{V}_{i})$;
		\STATE $\widehat{X}=\text{MHA}(\widehat{Q}, \widehat{K}, \widehat{V}) \in \mathbb{R}^{\frac{H}{\lambda}\times W\times\eta}$\\
		\ \ \ \ \ $=\text{Concat}(head_1,...,head_n)W^\text{M}$;
		\STATE $X^{\text{SW}}=\text{Flatten}(\widehat{X})$;
		\STATE \textbf{return} $S^\text{O} \in \mathbb{R}^{\frac{H}{\lambda}\times W\eta}$
	\end{algorithmic}
\end{algorithm}
\subsection{Time Complexity Analysis}
\par As shown in Fig. \ref{fig-2} (a), MVAE adopts a two-channel VAE structure. Therefore, the time complexity of MVAE is $T_{\text{MVAE}}=T^{\text{I}}_{\text{encoder}}+T^{\text{I}}_{\text{decoder}}+T^{\text{T}}_{\text{encoder}}+T^{\text{T}}_{\text{decoder}}+T^{z}=O(4((H-1)\cdot 256(768+D))+D)\sim O(H-1)$. Subsequently, the time complexity of SW-MHA is $O(k(W^2\cdot \eta))$. Where $k$ denotes the length of the window sequence. $W$ denotes the window length. $\eta$ denotes the dimension of the input sequence features. Therefore, the time complexity of Stage 2 is $T_{\text{Satge 1}}=T_{\text{SW-Block}}+b\cdot T_{\text{W-Block}}$. where $T_{\text{SW-Block}}=O(2D\cdot W_1^2+4D(2D+2DW_1))\sim O(D\cdot (W_1)^2+D^2W_1)$, and $T_{\text{SW-Block}}=2DW_1\cdot (W_1)^2+( 4DW_1)^2\sim DW_1\cdot (H/\lambda_1)^2+(DW_1)^2$. Thus, $T_{\text{Satge 1}}=O(D\cdot (W_1)^2+D^2W_1)+O(b(DW_1\cdot (H/\lambda_1)^2+(DW_1)^2))\sim O((H/\lambda_1)^2)$. Similarly, the time complexity of Stage 3 is $O((H/(\lambda_1\cdot \lambda_2))^2)$. Therefore, the overall time complexity of MS$^2$former is $O(H-1)+O((H/\lambda_1)^2)+O((H/(\lambda_1\cdot \lambda_2))^2)\sim O((H/\lambda_1)^2)$. Compared with the traditional MHA-based Transformer architecture ($O(H^2)$), MS$^2$former is superior in terms of operational efficiency (see Table \ref{table-ab-memory}).

\begin{figure*}[h]
	\center{\includegraphics[width=1.03\linewidth]  {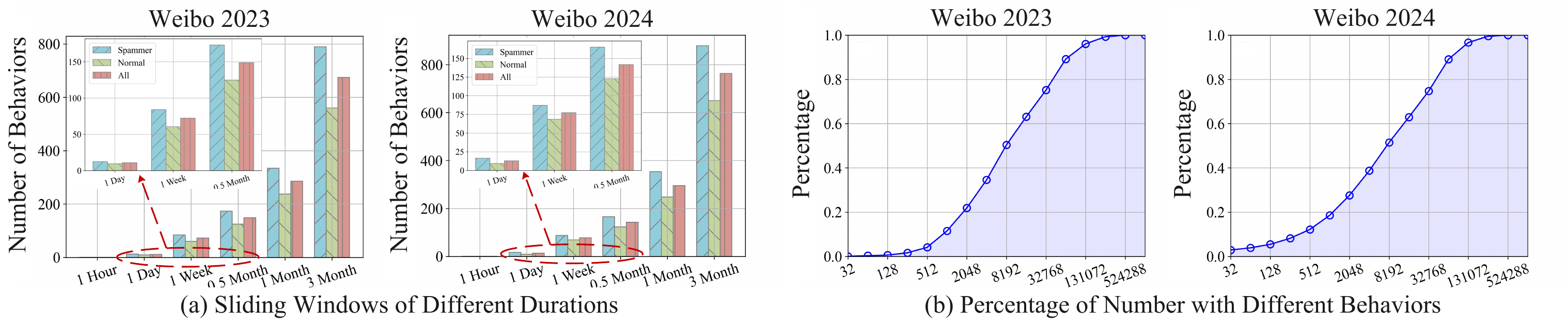}} 
	\caption{Statistics from two publicly available datasets.}
	\label{fig-Parameters-windows}
\end{figure*}
\begin{figure*}[h]
	\center{\includegraphics[width=1.03\linewidth]  {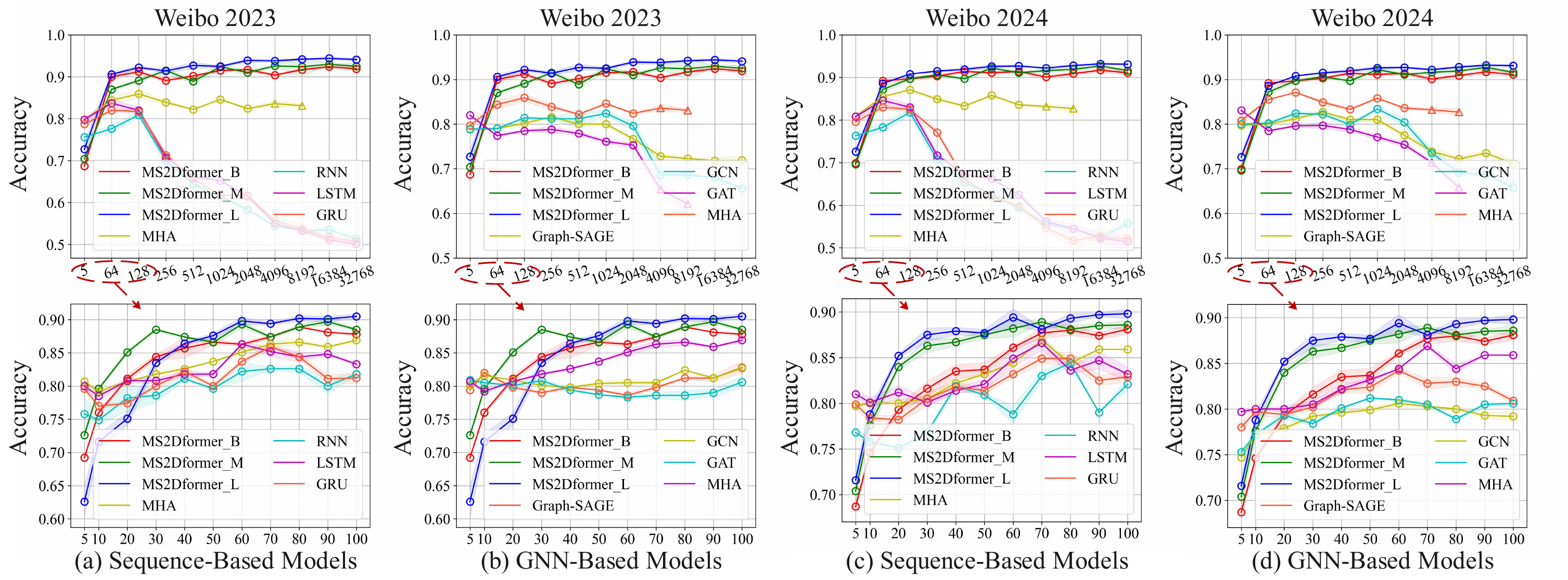}} 
	\caption{The influence of different numbers of historical behaviors on model training.}
	\label{fig-behavior}
\end{figure*}
\section{Experiment and Analysis}
\subsection{Experimental Data}	
\begin{table}[t]
	\renewcommand{\arraystretch}{1.3}
	\centering
	\caption{The statistics of the two datasets}
	\begin{tabular}{c|c c}
		\toprule[1.5pt]
		Statistic & Weibo 2023  & Weibo 2024 \\ \hline \hline
		$\#$ of Users& $684$ & $1971$ \\ \hline
		$\#$ of Normal& $343$ & $1019$ \\ \hline
		$\#$ of Spammer & $341$ & $952$ \\ \hline
		Avg. Behavior Length& $26,192$ & $25,310$ \\ \hline
		Max Behavior Length& $292,491$ & $308,798$ \\ \hline
		Min Behavior Length& $15$ & $1$ \\ \hline
		\bottomrule[1.5pt]
	\end{tabular}
	\label{table-datasets}
\end{table}
\par We conducted model training on two publicly available fake news detection datasets \cite{Yang2024model}, i.e., Weibo V1\footnote{\url{https://github.com/yzhouli/DDCA-Rumor-Detection/tree/main/MisDerdect.}} and Weibo V2\footnote{\url{https://github.com/yzhouli/SocialNet/tree/master/Weibo.}}. Spammers are usually created by users who frequently send out fake news. However, these users may not be aware of the facts or be interested in forwarding the news, so not every user who has sent fake news is a real spammer. Therefore, we hired graduate experts in various fields to determine further the users who posted fake news in the two datasets disclosed by \cite{Yang2024model} to filter out the real spammers. Subsequently, we constructed and collected two publicly available spammer detection benchmark datasets, Weibo 2023 and Weibo 2024. Weibo V1 contains 1,000 fake news and 948 potential users of spammers flagged by the Weibo platform between 2022 and 2023, and only 47 users who sent more than two fake news. Therefore, Weibo 2023 identifies only 342 spammers. Similarly, Weibo V2 contains 5,661 fake news and 5,653 potential users of spammers that were flagged by the Weibo platform between 2022 and 2024. Therefore, Weibo 2023 identifies only 952 spammers. The statistics of the two datasets are shown in Table \ref{table-datasets} and Fig.\ref{fig-Parameters-windows}.
\subsection{Parameters Setting}
\par \textbf{Model Variants:} Three variants of the MS$^2$Dformer model are constructed to cope with different usage environments:
\begin{equation}
	\begin{split}
		\text{MS$^2$Dformer}\_\text{B}:\{D&=16,W=\{64,64\},\\\lambda&=\{32, 4\},B=\{3, 3\}\}\\
	\end{split}
	\label{eq-base}
\end{equation}
\begin{equation}
	\begin{split}
		\text{MS$^2$Dformer}\_\text{M}:\{D&=16,W=\{128,64\},\\\lambda&=\{32, 4\},B=\{3, 11\}\}\\
	\end{split}
\end{equation}
\begin{equation}
	\begin{split}
		\text{MS$^2$Dformer}\_\text{L}:\{D&=64,W=\{128 ,64\},\\\lambda&=\{32, 4\},B=\{7, 17\}\}
	\end{split}
\end{equation}
where $D$ represents the embedding dimension of the encoder in MVAE. $W$ represents the SW-MHA window size in Stage 2-3. $\lambda$ represents the window sliding step. $B$ represents the number of SW-MHA or W-MHA Transformer Block.
\par
\begin{figure}[t]
	\center{\includegraphics[width=1.03\linewidth]  {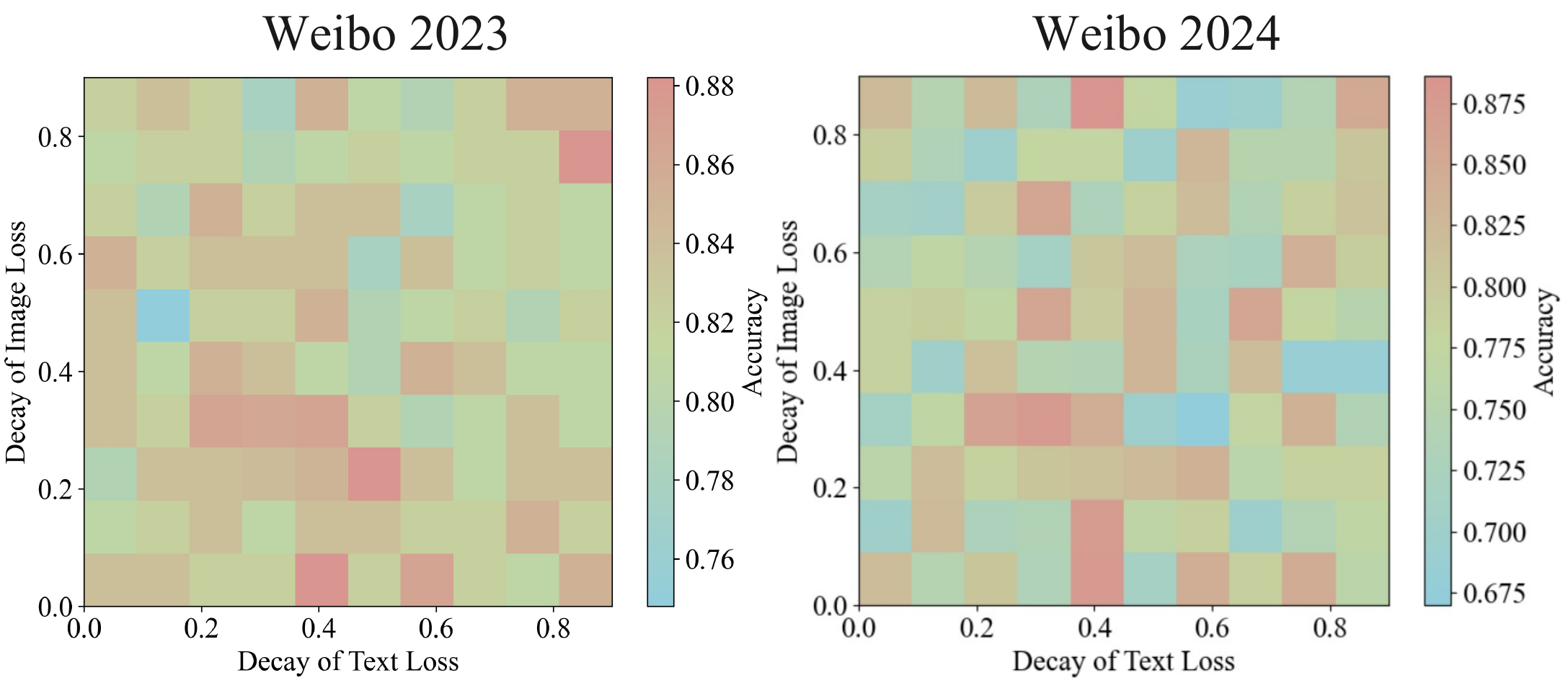}} 
	\caption{The influence of different values of $\psi_2$ and $\psi_3$ ($\psi_1=1$).}
	\label{fig-Parameters-decay}
\end{figure}

 \textbf{Hyper-parameter Settings:} We have written the model source code on the software platform of Tensorflow 2.9.0 and Python 3.8. Moreover, the dataset is divided into training, validation, and testing sets in the ratio of 7:2:1. Meanwhile, we used Adam as the optimizer of the model and set the learning rate to $1 \times 10^{-4}$. The model training time is 60 epochs, while the early stopping mechanism is introduced. Subsequently, we set the window size $W=64/128$ based on the average length of information dissemination (see Fig. \ref{fig-Parameters-windows} (a)). Afterward, the value of the multi-modal information decay factor of the loss function is verified using the control variable method (see Fig. \ref{fig-Parameters-decay}). In particular, the core of this work is to detect spammers, so $\psi_1$ is set to 1. It can be found that the best results are obtained when $\psi_2=0.3/0.3$ and $\psi_3=0.4/0.4$.
\par \textbf{Baseline Settings:} We selected advanced baseline algorithms in three dimensions. Firstly, in terms of algorithms, we selected traditional deep learning algorithms, including GCN\cite{li2019spam}, GAT\cite{zhang2023detecting}, Graph-SAGE\cite{zhang2024predicting}, RNN\cite{zhang2023rumor}, LSTM\cite{babu2023efficient}, GRU\cite{GRU}, and MHA\cite{rao2023hybrid}. Secondly, we selected Graph-U-Nets\cite{gao2019graph}, R-GCN\cite{generale2022scaling}, and ChebNet\cite{he2022convolutional} on the basis of GCN. Finally, from the task dimension, we chose specialized spammer detection models, i.e., MDGCN\cite{deng2023markov}, Graph Transformer\cite{GraphTrans}, and Adver-GCN\cite{zhang2022detecting}.
\subsection{Validity Analysis of Historical Behavioral Length}
\begin{table*}[htbp]
	\renewcommand{\arraystretch}{1.3}
	\caption{Comparison of the performance for different models on the Weibo 2023 dataset}
	\label{table_acc_v1}
	\centering	
	\begin{tabular}{cccccccc}
		\toprule[1.5pt]
		\multicolumn{1}{c|}{\multirow{2}*{Method}}&\multicolumn{1}{c|}{\multirow{2}*{Accuracy}}&\multicolumn{3}{c|}{Normal}&\multicolumn{3}{c}{Spammer}\\
		\cline{3-5} \cline{6-8}
		\multicolumn{1}{c|}{}&\multicolumn{1}{c|}{}&\multicolumn{1}{c}{\multirow{1}*{Precision}}&\multicolumn{1}{c}{Recall}&\multicolumn{1}{c}{F1}&\multicolumn{1}{c}{\multirow{1}*{Precision}}&\multicolumn{1}{c}{Recall}&\multicolumn{1}{c}{F1}\\
		\hline \hline
		GAT & $0.816_{\pm 0.007}$ & $0.83_{\pm 0.003}$ & $0.794_{\pm 0.015}$ & $0.812_{\pm 0.009}$ & $0.803_{\pm 0.011}$ & $0.838_{\pm 0.001}$ & $0.820_{\pm 0.006}$\\
		Graph-SAGE & $0.828_{\pm 0.003}$ & $0.821_{\pm 0.003}$ & $0.838_{\pm 0.014}$ & $0.830_{\pm 0.005}$ & $0.835_{\pm 0.011}$ & $0.817_{\pm 0.008}$ & $0.826_{\pm 0.002}$\\
		GCN & $0.832_{\pm 0.007}$ & $0.798_{\pm 0.001}$ & $0.890_{\pm 0.022}$ & $0.842_{\pm 0.010}$ & $0.874_{\pm 0.020}$ & $0.773_{\pm 0.008}$ & $0.820_{\pm 0.005}$\\
		Graph-U-Nets & $0.758_{\pm 0.008}$ & $0.788_{\pm 0.056}$ & $0.717_{\pm 0.103}$ & $0.743_{\pm 0.030}$ & $0.749_{\pm 0.044}$ & $0.800_{\pm 0.110}$ & $0.767_{\pm 0.036}$\\ 
		R-GCN & $0.820_{\pm 0.004}$ & $0.778_{\pm 0.019}$ & $0.897_{\pm 0.029}$ & $0.832_{\pm 0.002}$ & $0.881_{\pm 0.026}$ & $0.742_{\pm 0.037}$ & $0.804_{\pm 0.011}$\\
		MDGCN & $0.842_{\pm 0.007}$ & $0.840_{\pm 0.034}$ & $0.849_{\pm 0.037}$ & $0.843_{\pm 0.001}$ & $0.848_{\pm 0.020}$ & $0.834_{\pm 0.051}$ & $0.84_{\pm 0.016}$\\  
		ChebNet & $0.834_{\pm 0.001}$ & $0.826_{\pm 0.005}$ & $0.849_{\pm 0.007}$ & $0.837_{\pm 0.001}$ & $0.844_{\pm 0.005}$ & $0.820_{\pm 0.008}$ & $0.832_{\pm 0.001}$\\
		Adver-GCN & $0.844_{\pm 0.007}$ & $0.832_{\pm 0.016}$ & $0.866_{\pm 0.007}$ & $0.848_{\pm 0.004}$ & $0.858_{\pm 0.002}$ & $0.822_{\pm 0.022}$ & $0.840_{\pm 0.010}$\\ 
		Graph Transformer & $0.864_{\pm 0.007}$ & $0.899_{\pm 0.008}$ & $0.825_{\pm 0.021}$ & $0.862_{\pm 0.008}$ & $0.832_{\pm 0.015}$ & $0.904_{\pm 0.007}$ & $0.867_{\pm 0.007}$\\\hline \hline
		
		RNN &  $0.832_{\pm 0.022}$ & $0.816_{\pm 0.005}$ & $0.852_{\pm 0.052}$ & $0.834_{\pm 0.027}$ & $0.850_{\pm 0.043}$ & $0.810_{\pm 0.008}$ & $0.829_{\pm 0.017}$\\
		GRU & $0.863_{\pm 0.011}$ & $0.865_{\pm 0.003}$ & $0.858_{\pm 0.023}$ & $0.861_{\pm 0.013}$ & $0.862_{\pm 0.019}$ & $0.868_{\pm 0.001}$ & $0.864_{\pm 0.010}$\\
		LSTM & $0.866_{\pm 0.008}$ & $0.884_{\pm 0.020}$ & $0.843_{\pm 0.008}$ & $0.862_{\pm 0.006}$ & $0.852_{\pm 0.003}$ & $0.890_{\pm 0.022}$ & $0.870_{\pm 0.009}$\\ 
		MHA & $0.872_{\pm 0.007}$ & $0.871_{\pm 0.012}$ & $0.871_{\pm 0.001}$ & $0.871_{\pm 0.006}$ & $0.874_{\pm 0.003}$ & $0.874_{\pm 0.015}$ & $0.874_{\pm 0.009}$\\ \hline \hline
		
		MS$^2$Dformer\_B & $0.923_{\pm 0.004}$ & $0.913_{\pm 0.001}$ & $0.932_{\pm 0.007}$ & $0.923_{\pm 0.004}$ & $0.932_{\pm 0.007}$ & $\underline{0.912}_{\pm 0.001}$ & $0.922_{\pm 0.003}$\\
		MS$^2$Dformer\_M & $\underline{0.931}_{\pm 0.004}$ & $\underline{0.914}_{\pm 0.011}$ & $\underline{0.948}_{\pm 0.022}$ & $\underline{0.931}_{\pm 0.005}$ & $\underline{0.947}_{\pm 0.021}$ & $0.912_{\pm 0.015}$ & $\underline{0.929}_{\pm 0.003}$\\
		MS$^2$Dformer\_L & $\textbf{0.941}_{\pm 0.004}$ & $\textbf{0.926}_{\pm 0.018}$ & $\textbf{0.959}_{\pm 0.015}$ & $\textbf{0.942}_{\pm 0.002}$ & $\textbf{0.958}_{\pm 0.014}$ & $\textbf{0.923}_{\pm 0.022}$ & $\textbf{0.940}_{\pm 0.005}$\\ \hline
		\bottomrule[1.5pt]
	\end{tabular}
\end{table*}
\par In social platforms, the distribution of the number of historical behaviors of users varies widely. As shown in Fig. \ref{fig-Parameters-windows} (b), the number of historical behaviors of 16,384/32,768 is the inflection point of this distribution, with more than 60/70\% of users exceeding this threshold. Therefore, this range of historical behaviors is verified for the model performance (shown in Fig. \ref{fig-behavior}). In particular, the performance of the GNN-based models, i.e., GCN, GAT, and Graph-SAGE, decreases dramatically when the threshold of 2048, the number of behaviors, is exceeded. At this point, the spread graph constructed by the user's historical behaviors overlaps, and the traditional GNN models cannot effectively mine the global and local structural information. On the other hand, the performance of models based on sequence modeling, i.e., RNN, LSTM, and GRU, decreases dramatically when exceeding the threshold of 128 as the number of behaviors. This is because such sequence models do not effectively measure ultra-long-term and short-term interactions when confronted with ultra-long sequences. Finally, a similar exercise was conducted to examine the performance of the traditional Multi-head Attention Mechanism (MHA) when faced with different lengths of historical behavioral sequences. It can be found that MHA can effectively measure long- and short-term interactions. However, due to the memory pressure of $QK^{\text{T}}$, MHA cannot run when facing ultra-long sequences. Specifically, $\circ$ and $\triangle$ in Fig. \ref{fig-behavior} (a) and (c) represent that the training platform is RTX 4060 (16 GB) or A800 (80 GB).
\begin{table}[htbp]
	\renewcommand{\arraystretch}{1.3}
	\centering	
	\caption{Performance comparison of different multi-modal fusion strategies}
	\resizebox{0.48\textwidth}{!}{
		\begin{tabular}{ccccccc}
			\toprule[1.5pt]
			\multicolumn{1}{c|}{\multirow{2}*{Method}}&\multicolumn{3}{c|}{\multirow{1}*{Weibo 2023}}&\multicolumn{3}{c}{Weibo 2024}\\
			\cline{2-4} \cline{5-7}
			\multicolumn{1}{c|}{}&\multicolumn{1}{c}{Base}&\multicolumn{1}{c}{Middle}&\multicolumn{1}{c|}{Large}&\multicolumn{1}{c}{Base}&\multicolumn{1}{c}{Middle}&\multicolumn{1}{c}{Large}\\
			\hline \hline
			Our (MVAE \& (Bert+ViT)) & $\textbf{0.923}$ & $\textbf{0.931}$ & $\textbf{0.941}$ & $\underline{0.917}$ & $\textbf{0.926}$ & $\textbf{0.935}$ \\ \hline \hline
			-w Text \& w/o Image & $0.859$ & $0.863$ & $0.861$ & $0.863$ & $0.863$ & $0.881$ \\
			-w Image \& w/o Text & $0.904$ & $0.889$ & $0.897$ & $0.875$ & $0.884$ & $0.862$ \\
			-w Add Image \& Text & $0.889$ & $0.904$ & $0.924$ & $0.889$ & $0.897$ & $0.901$ \\ \hline \hline
			-w CLIP \& w/o (Bert+ViT) & $0.909$ & $0.914$ & $0.933$ & $0.913$ & $0.914$ & $0.928$ \\
			-w ALIGN \& w/o (Bert+ViT) & $0.915$ & $0.923$ & $\underline{0.937}$ & $\textbf{0.917}$ & $0.921$ & $0.928$ \\
			-w BLIP-2 \& w/o (Bert+ViT)  & $\underline{0.917}$ & $\underline{0.928}$ & $0.935$ & $0.916$ & $\underline{0.924}$ & $\underline{0.929}$ \\ 
			\hline
			\bottomrule[1.5pt]
	\end{tabular}}
	\label{table-ab-multi-model}
\end{table}
\begin{table*}[htbp]
	\renewcommand{\arraystretch}{1.3}
	\caption{Comparison of the performance for different models on the Weibo 2024 dataset}
	\label{table_acc_v2}
	\centering	
	\begin{tabular}{cccccccc}
		\toprule[1.5pt]
		\multicolumn{1}{c|}{\multirow{2}*{Method}}&\multicolumn{1}{c|}{\multirow{2}*{Accuracy}}&\multicolumn{3}{c|}{Normal}&\multicolumn{3}{c}{Spammer}\\
		\cline{3-5} \cline{6-8}
		\multicolumn{1}{c|}{}&\multicolumn{1}{c|}{}&\multicolumn{1}{c}{\multirow{1}*{Precision}}&\multicolumn{1}{c}{Recall}&\multicolumn{1}{c}{F1}&\multicolumn{1}{c}{\multirow{1}*{Precision}}&\multicolumn{1}{c}{Recall}&\multicolumn{1}{c}{F1}\\
		\hline \hline
		GAT & $0.810_{\pm 0.003}$ & $0.803_{\pm 0.018}$ & $0.840_{\pm 0.034}$ & $0.820_{\pm 0.008}$ & $0.821_{\pm 0.024}$ & $0.779_{\pm 0.034}$ & $0.799_{\pm 0.006}$\\
		Graph-SAGE & $0.842_{\pm 0.005}$ & $0.817_{\pm 0.014}$ & $0.894_{\pm 0.037}$ & $0.853_{\pm 0.009}$ & $0.876_{\pm 0.035}$ & $0.786_{\pm 0.029}$ & $0.828_{\pm 0.005}$\\ 
		GCN & $0.806_{\pm 0.005}$ & $0.823_{\pm 0.029}$ & $0.796_{\pm 0.042}$ & $0.808_{\pm 0.010}$ & $0.790_{\pm 0.025}$ & $0.816_{\pm 0.044}$ & $0.802_{\pm 0.009}$\\
		Graph-U-Nets & $0.744_{\pm 0.002}$ & $0.804_{\pm 0.023}$ & $0.664_{\pm 0.035}$ & $0.726_{\pm 0.011}$ & $0.702_{\pm 0.011}$ & $0.831_{\pm 0.037}$ & $0.761_{\pm 0.01}$\\
		R-GCN & $0.842_{\pm 0.004}$ & $0.832_{\pm 0.031}$ & $0.871_{\pm 0.047}$ & $0.85_{\pm 0.006}$ & $0.856_{\pm 0.038}$ & $0.811_{\pm 0.052}$ & $0.832_{\pm 0.011}$\\
		MDGCN & $0.841_{\pm 0.003}$ & $0.85_{\pm 0.012}$ & $0.842_{\pm 0.012}$ & $0.846_{\pm 0.001}$ & $0.831_{\pm 0.007}$ & $0.838_{\pm 0.018}$ & $0.835_{\pm 0.006}$\\ 
		ChebNet & $0.812_{\pm 0.004}$ & $0.838_{\pm 0.048}$ & $0.798_{\pm 0.062}$ & $0.814_{\pm 0.010}$ & $0.798_{\pm 0.034}$ & $0.828_{\pm 0.074}$ & $0.808_{\pm 0.017}$\\
		Adver-GCN & $0.851_{\pm 0.004}$ & $0.864_{\pm 0.021}$ & $0.832_{\pm 0.022}$ & $0.848_{\pm 0.002}$ & $0.840_{\pm 0.012}$ & $0.870_{\pm 0.029}$ & $0.854_{\pm 0.008}$\\ 
		Graph Transformer & $0.865_{\pm 0.008}$ & $0.898_{\pm 0.006}$ & $0.827_{\pm 0.023}$ & $0.861_{\pm 0.009}$ & $0.838_{\pm 0.017}$ & $0.902_{\pm 0.008}$ & $0.869_{\pm 0.006}$\\ \hline \hline
		RNN & $0.838_{\pm 0.004}$ & $0.839_{\pm 0.04}$ & $0.840_{\pm 0.052}$ & $0.837_{\pm 0.006}$ & $0.846_{\pm 0.033}$ & $0.836_{\pm 0.058}$ & $0.838_{\pm 0.013}$\\ 
		GRU & $0.856_{\pm 0.008}$ & $0.866_{\pm 0.003}$ & $0.840_{\pm 0.022}$ & $0.853_{\pm 0.010}$ & $0.848_{\pm 0.017}$ & $0.872_{\pm 0.008}$ & $0.860_{\pm 0.005}$\\
		LSTM & $0.875_{\pm 0.004}$ & $0.883_{\pm 0.016}$ & $0.863_{\pm 0.03}$ & $0.872_{\pm 0.008}$ & $0.869_{\pm 0.022}$ & $0.887_{\pm 0.022}$ & $0.878_{\pm 0.001}$\\ 
		MHA & $0.883_{\pm 0.004}$ & $0.885_{\pm 0.007}$ & $0.878_{\pm 0.003}$ & $0.881_{\pm 0.003}$ & $0.881_{\pm 0.001}$ & $0.887_{\pm 0.007}$ & $0.884_{\pm 0.004}$\\ \hline \hline
		
		MS$^2$Dformer\_B & $0.917_{\pm 0.004}$ & $\underline{0.928}_{\pm 0.013}$ & $0.903_{\pm 0.022}$ & $0.915_{\pm 0.006}$ & $0.907_{\pm 0.018}$ & $\underline{0.930}_{\pm 0.014}$ & $0.919_{\pm 0.004}$\\
		MS$^2$Dformer\_M & $\underline{0.926}_{\pm 0.007}$ & $0.899_{\pm 0.012}$ & $\textbf{0.959}_{\pm 0.015}$ & $\underline{0.928}_{\pm 0.007}$ & $\underline{0.917}_{\pm 0.015}$ & $0.893_{\pm 0.015}$ & $\underline{0.924}_{\pm 0.008}$\\
		MS$^2$Dformer\_L & $\textbf{0.935}_{\pm 0.004}$ & $\textbf{0.937}_{\pm 0.001}$ & $\underline{0.930}_{\pm 0.008}$ & $\textbf{0.933}_{\pm 0.004}$ & $\textbf{0.932}_{\pm 0.007}$ & $\textbf{0.938}_{\pm 0.004}$ & $\textbf{0.935}_{\pm 0.003}$\\ \hline
		\bottomrule[1.5pt]
	\end{tabular}
\end{table*}
\begin{figure}[t]
	\center{\includegraphics[width=1\linewidth]  {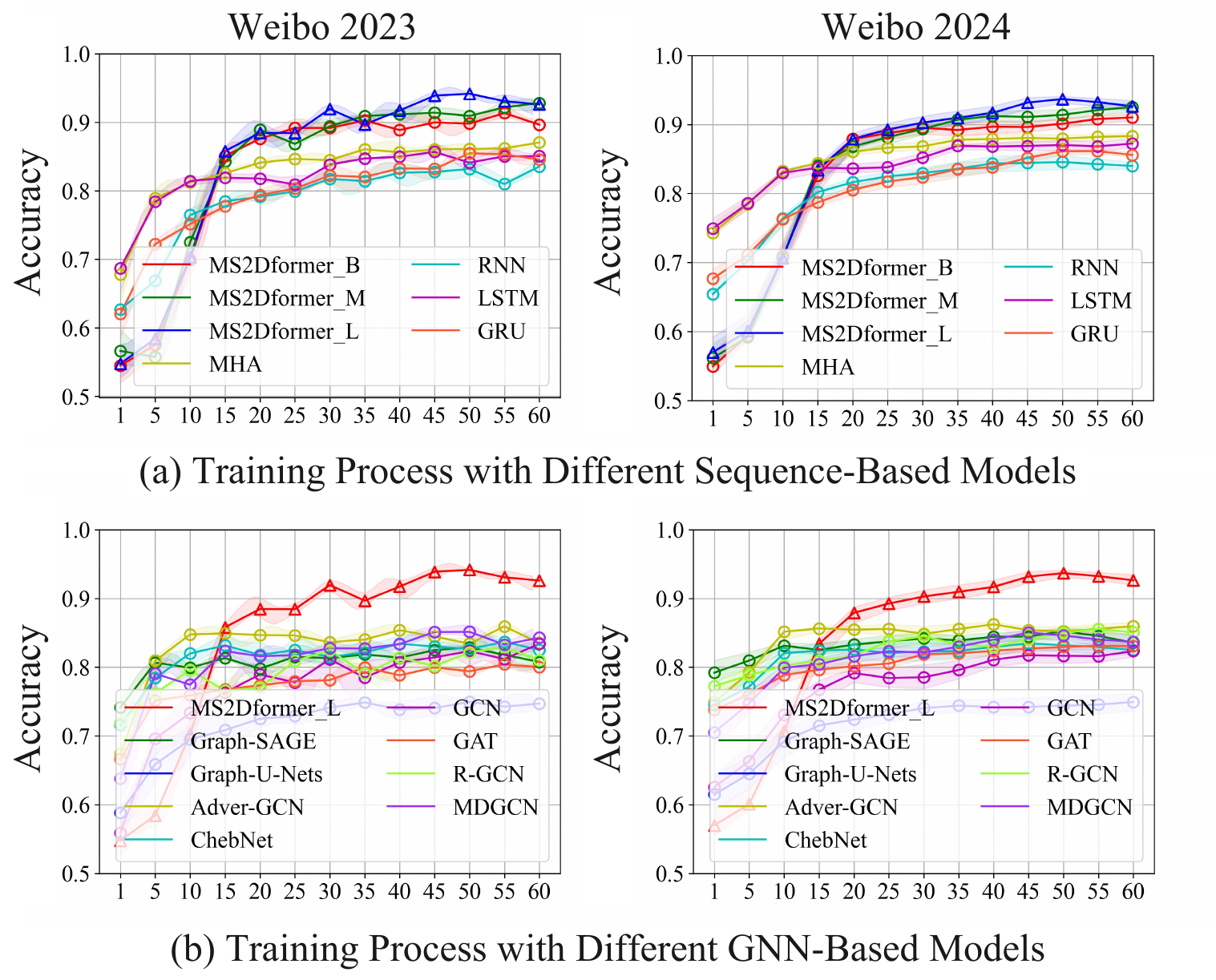}} 
	\caption{Comparison of model training based on optimal parameter settings.}
	\label{fig-models}
\end{figure}
\begin{table*}[t]
	\renewcommand{\arraystretch}{1.3}
	\centering	
	\caption{GPU  platform operation with different MHA mechanisms (behavior size =16384). SPU represents the number of \textbf{s}econds that the model consumes to \textbf{p}rocess a individual \textbf{u}ser.}
		\begin{tabular}{ccccccccc}
			\toprule[1.5pt]
			\multicolumn{1}{c|}{\multirow{2}*{Method}}&\multicolumn{1}{c|}{\multirow{2}*{GPU Platform }}&\multicolumn{1}{c|}{\multirow{2}*{Run}}&\multicolumn{2}{c|}{\multirow{1}*{Base}}&\multicolumn{2}{c|}{\multirow{1}*{Middle}}&\multicolumn{2}{c}{\multirow{1}*{Large}}\\
			\cline{4-9}
			\multicolumn{1}{c|}{}&\multicolumn{1}{c|}{}&\multicolumn{1}{c|}{}&\multicolumn{1}{c}{Params}&\multicolumn{1}{c|}{SPU}&\multicolumn{1}{c}{Params}&\multicolumn{1}{c|}{SPU}&\multicolumn{1}{c}{Params}&\multicolumn{1}{c}{SPU}\\
			\hline \hline
			Our (SW-MHA \& W-MHA) & RTX 4060 (16 GB) & $\surd$ & $2.21$ M & $0.29$ s & $3.66$ M & $0.32$ s & $53.8$ M & $0.35$ s \\ \hline \hline
			-w CNN \& w/o SW-MHA & RTX 4060 (16 GB) & $\surd$ & $1.97$ M & $0.20$ s & $2.53$ M & $0.23$ s & $20.9$ M & $0.27$ s \\ \hline \hline
			-w MHA \& w MVAE & A800 (80 GB) & $\times$ & $-$ & $-$ & $-$ & $-$ & $-$ & $-$  \\ \hline \hline
			-w SW-SMHA \& w MVAE & V100 (32 GB) & $\surd$ & $1.73$ M & $77.2$ s & $1.82$ M & $172.6$ s & $6.41$ M & $181.3$ s \\
			-w BRSW-SMHA \& w MVAE & V100 (32 GB) & $\surd$ & $1.73$ M & $0.87$ s & $1.82$ M & $1.61$ s & $6.41$ M & $3.20$ s \\
			-w BSW-SMHA \& w MVAE & V100 (32 GB) & $\surd$ & $1.73$ M & $1.05$ s & $1.82$ M & $1.22$ s & $6.41$ M & $3.78$ s \\
			-w BDW-SMHA \& w MVAE & V100 (32 GB) & $\surd$ & $1.73$ M & $1.55$ s & $1.82$ M & $1.56$ s & $6.41$ M & $3.20$ s \\
			-w BGSW-SMHA \& w MVAE & V100 (32 GB) & $\surd$ & $1.73$ M & $0.89$ s & $1.82$ M & $1.74$ s & $6.41$ M & $3.82$ s \\ \hline
			\bottomrule[1.5pt]
	\end{tabular}
	\label{table-ab-memory}
\end{table*}
\subsection{Overall Performance Analysis}
\par The MS$^2$Dformer and baseline models are trained based on the optimal behavior length, the training process is shown in Fig. \ref{fig-models}, and the training results are shown in Table \ref{table_acc_v1}-\ref{table_acc_v2}. 
\par Comparing the GNN variants, the MS$^2$Dformer\_L model performance improves $+0.077$ and $+0.07$, and the MHA model performance improves $+0.008$ and $+0.018$. It can be found that the sequence modeling strategy is more suitable for the task at hand. Subsequently, comparing the sequence modeling variants, the MS$^2$Dformer\_L model performance is improved by $+0.069$ and $+0.052$. The MS$^2$Dformer architecture is able to efficiently mine the sequence features, thus validating its effectiveness.
\subsection{Ablation Study}
\par \textbf{Multi-Modal Fusion:} As shown in Table \ref{table-ab-multi-model}, compared to the uni-modal modeling strategies, i.e., -w Text \& w/o Image and -w Image \& w/o Text, the fusion feature strategy (MVAE \& (Bert+ViT)) improves the performance by $+3.7$\% to $+8.2$\%. Thus, the validity of the multi-modal fusion strategy can be demonstrated. Subsequently, comparing the direct fusion strategy (-w Add Image \& Text), there also exists a substantial performance improvement in the fusion strategy (-w CLIP\cite{yang2022chinese}/ALIGN\cite{jia2021scaling}/BLIP-2\cite{li2023blip} \& w/o (Bert+ViT)) using pre-trained models. For instance, the CLIP model constructs models based on an adversarial learning strategy. Subsequently, more generalized knowledge is learned over a wide range of datasets. Therefore, the three pre-trained CLIP/ALIGN/BLIP-2 models are used with better results. However, they could not focus more on the current task. Therefore, a combination strategy of MVAE \& (Bert+ViT) and full training was used to achieve the best performance.
\begin{table}[t]
	\renewcommand{\arraystretch}{1.3}
	\centering	
	\caption{Performance comparison of different MHA mechanisms}
	\resizebox{0.48\textwidth}{!}{
		\begin{tabular}{ccccccc}
			\toprule[1.5pt]
			\multicolumn{1}{c|}{\multirow{2}*{Method}}&\multicolumn{3}{c|}{\multirow{1}*{Weibo 2023}}&\multicolumn{3}{c}{Weibo 2024}\\
			\cline{2-4} \cline{5-7}
			\multicolumn{1}{c|}{}&\multicolumn{1}{c}{Base}&\multicolumn{1}{c}{Middle}&\multicolumn{1}{c|}{Large}&\multicolumn{1}{c}{Base}&\multicolumn{1}{c}{Middle}&\multicolumn{1}{c}{Large}\\
			\hline \hline
			Our (SW-MHA \& W-MHA) & $\textbf{0.923}$ & $\textbf{0.931}$ & $\textbf{0.941}$ & $\textbf{0.917}$ & $\textbf{0.926}$ & $\textbf{0.935}$ \\ \hline \hline
			-w CNN \& w/o SW-MHA &  $0.911$ & $0.912$ & $0.919$ & $0.912$ & $0.910$ & $0.914$ \\ \hline \hline
			-w SW-SMHA \& w MVAE & $\underline{0.919}$ & $0.924$ & $0.927$ & $0.915$ & $\underline{0.921}$ & $\underline{0.928}$ \\
			-w BRSW-SMHA \& w MVAE & $0.907$ & $0.912$ & $0.913$ & $0.907$ & $0.913$ & $0.911$ \\
			-w BSW-SMHA \& w MVAE & $0.913$ & $0.918$ & $0.922$ & $0.913$ & $0.918$ & $0.918$ \\
			-w BDW-SMHA \& w MVAE & $0.912$ & $0.921$ & $0.923$ & $0.901$ & $0.919$ & $0.908$ \\
			-w BGSW-SMHA \& w MVAE & $0.918$ & $0.926$ & $\underline{0.929}$ & $\underline{0.916}$ & $0.917$ & $0.927$ \\ \hline
			\bottomrule[1.5pt]
	\end{tabular}}
	\label{table-ab-MHA}
\end{table}
\par \textbf{Sequence Modeling Variants:} This work proposes a new MHA mechanism based on hierarchical split windows. Subsequently, the modified MHA mechanism can run on GPU platforms with lower memory (see Table \ref{table-ab-memory} up). Currently, sparse attention mechanisms are proposed in academia to solve the problem of attention computation for ultra-long sequences. Sparse attention has made excellent contributions in relieving GPU memory pressure (see Table \ref{table-ab-memory} bottom). However, sparse attention inherently fails to effectively mine the long-term interactions of ultra-long sequences. Thus, it performs poorly in the current task (see Table \ref{table-ab-MHA}). In particular, to address the problem of modeling ultra-long sequences, CNN is introduced before the Transformer block, thus reducing the input sequence to a length supported by the GPU platform. The SW-MHA mechanism proposed in this work works similarly to the CNN. However, the CNN cannot quantify short-term interactions efficiently, thus performing less (see Table \ref{table-ab-MHA}). Finally, as shown in Table \ref{table-ab-memory}, our proposed model completes model training with more than 53 million parameters on the RTX 4060 (16 GB) platform with the smallest memory. Moreover, the model has a low SPU. Thus, the validity of our proposed model is validated once again.
\begin{table}[htbp]
	\renewcommand{\arraystretch}{1.3}
	\centering	
	\caption{Performance comparison of MS$^2$Dformer\_B models based on different position encoding}
	\resizebox{0.3\textwidth}{!}{
	\begin{tabular}{ccc}
		\toprule[1.5pt]
		Method & Weibo 2023 & Weibo 2024\\
		\hline \hline
		Our (APE \& APE) & $\textbf{0.923}$ & $\textbf{0.907}$\\ \hline \hline
		-w/o PE \& w/o PE & $0.896$ & $0.891$ \\
		-w/o PE \& w APE & $0.901$ & $0.897$ \\
		-w/o PE \& w TPE & $0.915$ & $0.899$ \\ \hline \hline
		-w APE \& w/o PE & $0.911$ & $0.901$ \\
		-w APE \& w TPE & $\underline{0.919}$ & $\underline{0.905}$ \\ \hline \hline
		-w TPE \& w/o PE & $0.896$ & $0.887$ \\
		-w TPE \& w APE & $0.909$ & $0.902$ \\
		-w TPE \& w TPE & $0.898$ & $0.903$ \\ \hline
		\bottomrule[1.5pt]
	\end{tabular}}
	\label{table-ab-PE}
\end{table}
\par \textbf{Position Encoding:} The spammer detection model based on sequence modeling focuses on the long-term and short-term interaction features of historical behavioral sequences in terms of temporal features. In the MHA mechanism, position encoding aims to enable the model to know the positional relationship of the sequences. Therefore, the model can decline dramatically when removing the position encoding (PE) for SW-MHA and W-MHA (see Table \ref{table-ab-PE}). When combined with absolute position encoding (APE), the model can perceive temporal features effectively. The model may acquire more effective position information when using trainable position encoding (TPE). However, TPE may also disrupt the time dimension's long- and short-term relationships. Therefore, the performance of models using the TPE mechanism is unstable. In summary, the model works best when APE is used for the SW-MHA and W-MHA.
\begin{figure}[t]
	\center{\includegraphics[width=1\linewidth]  {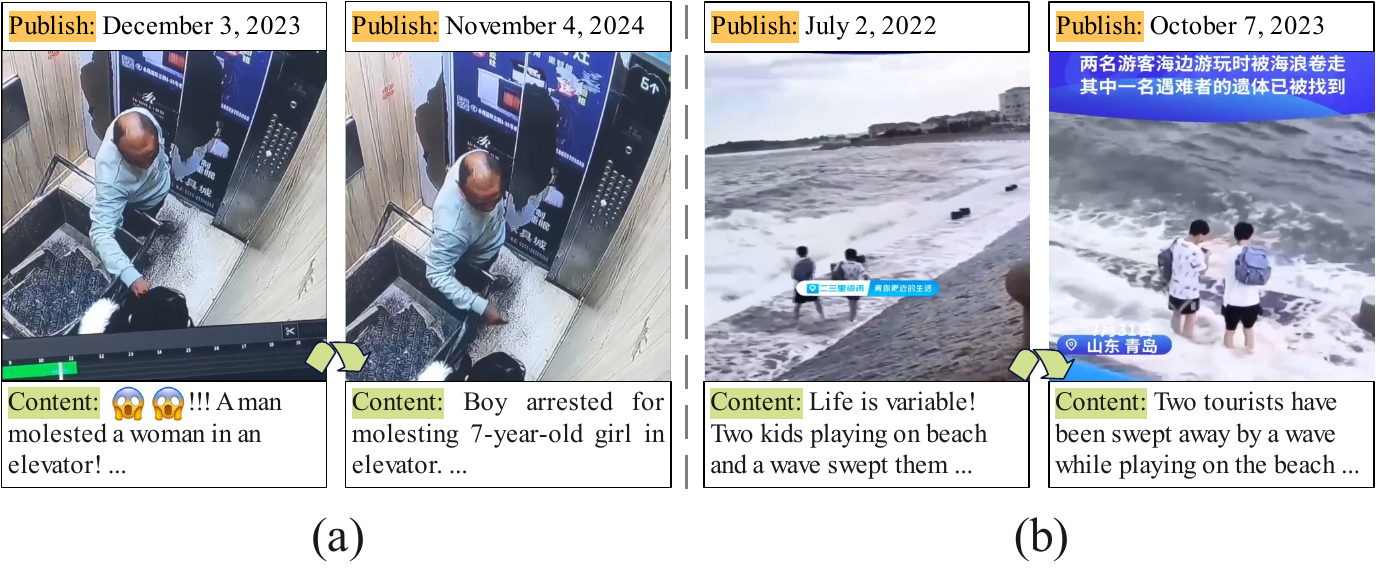}} 
	\caption{The case of ultra-long history interactions with similar user behavior. (a) and (b) are selected from different historical behaviors of a individual user.}
	\label{fig-ultra-long-term}
\end{figure}
\subsection{Case Study}
\par In social platforms, users usually post information with noise affecting the model to extract meaningful features (see Fig. \ref{fig-inspire}). Using MVAE, our proposed model can effectively reduce the noise influence (see Table \ref{table-ab-multi-model}). Meanwhile, users have short-term self-behavioral interactions. Therefore, we adopt the strategy of split window to mine the short-term interaction relations (see Fig. \ref{fig-Parameters-windows}). However, because online social has been developed for many years, spammers often guide public opinion again with the help of already published or outdated information (see Fig. \ref{fig-ultra-long-term}). This interaction of ultra-long-term behavior is important for identifying latent spammers. Traditional MHA is affected by explicit memory, so it cannot handle ultra-long-term historical interactions (see Table \ref{table-ab-memory} up). To address this problem, scholars have proposed SMHA to model ultra-long sequences. However, none of them can avoid constructing the $QK^\text{T}$ matrix (see Fig. \ref{fig-MHAs}). Finally, we propose a hierarchical attention mechanism based on a split window according to MHA and SMHA, which exists a considerable performance improvement (see Table \ref{table_acc_v1}-\ref{table_acc_v2}), arithmetic consumption, SPU, and parameter size (see Table \ref{table-ab-memory}).
\section{Conclusions}
\par In this paper, we propose a model for multi-modal sequential spammer detection. To relieve the effect of multi-modal noise, a two-channel VAE mechanism is first constructed to complete the history behavior Tokenization. Subsequently, to model ultra-long historical behavior sequences, a hierarchical multi-head attention mechanism based on the split window is proposed for the first time. The MS$^2$Dformer architecture, with the largest parameters, lower processing time, and huge performance improvement, is trained in a low-computing power platform (RTX 4060 (16 GB)). In the future, the performance of the hierarchical multi-head attention mechanism will be tested for other ultra-long sequence representation tasks, such as recommendation systems.


\begin{thebibliography}{10}
\providecommand{\url}[1]{#1}
\csname url@samestyle\endcsname
\providecommand{\newblock}{\relax}
\providecommand{\bibinfo}[2]{#2}
\providecommand{\BIBentrySTDinterwordspacing}{\spaceskip=0pt\relax}
\providecommand{\BIBentryALTinterwordstretchfactor}{4}
\providecommand{\BIBentryALTinterwordspacing}{\spaceskip=\fontdimen2\font plus
\BIBentryALTinterwordstretchfactor\fontdimen3\font minus
  \fontdimen4\font\relax}
\providecommand{\BIBforeignlanguage}[2]{{%
\expandafter\ifx\csname l@#1\endcsname\relax
\typeout{** WARNING: IEEEtran.bst: No hyphenation pattern has been}%
\typeout{** loaded for the language `#1'. Using the pattern for}%
\typeout{** the default language instead.}%
\else
\language=\csname l@#1\endcsname
\fi
#2}}
\providecommand{\BIBdecl}{\relax}
\BIBdecl

\bibitem{devlin2018bert}
J.~Devlin, ``Bert: Pre-training of deep bidirectional transformers for language
  understanding,'' \emph{arXiv preprint arXiv:1810.04805}, 2018.

\bibitem{radford2019language}
A.~Radford, J.~Wu, R.~Child, D.~Luan, D.~Amodei, I.~Sutskever \emph{et~al.},
  ``Language models are unsupervised multitask learners,'' \emph{OpenAI blog},
  vol.~1, no.~8, p.~9, 2019.

\bibitem{brown2020language}
T.~B. Brown, ``Language models are few-shot learners,'' \emph{arXiv preprint
  arXiv:2005.14165}, 2020.

\bibitem{beltagy2020longformer}
I.~Beltagy, M.~E. Peters, and A.~Cohan, ``Longformer: The long-document
  transformer,'' \emph{arXiv preprint arXiv:2004.05150}, 2020.

\bibitem{achiam2023gpt}
J.~Achiam, S.~Adler, S.~Agarwal, L.~Ahmad, I.~Akkaya, F.~L. Aleman, D.~Almeida,
  J.~Altenschmidt, S.~Altman, S.~Anadkat \emph{et~al.}, ``Gpt-4 technical
  report,'' \emph{arXiv preprint arXiv:2303.08774}, 2023.

\bibitem{wang2024utilizing}
J.~Wang, J.~X. Huang, X.~Tu, J.~Wang, A.~J. Huang, M.~T.~R. Laskar, and
  A.~Bhuiyan, ``Utilizing bert for information retrieval: Survey, applications,
  resources, and challenges,'' \emph{ACM Computing Surveys}, vol.~56, no.~7,
  pp. 1--33, 2024.

\bibitem{liu2021swin}
Z.~Liu, Y.~Lin, Y.~Cao, H.~Hu, Y.~Wei, Z.~Zhang, S.~Lin, and B.~Guo, ``Swin
  transformer: Hierarchical vision transformer using shifted windows,'' in
  \emph{Proceedings of the IEEE/CVF international conference on computer
  vision}, 2021, pp. 10\,012--10\,022.

\bibitem{han2022survey}
K.~Han, Y.~Wang, H.~Chen, X.~Chen, J.~Guo, Z.~Liu, Y.~Tang, A.~Xiao, C.~Xu,
  Y.~Xu \emph{et~al.}, ``A survey on vision transformer,'' \emph{IEEE
  transactions on pattern analysis and machine intelligence}, vol.~45, no.~1,
  pp. 87--110, 2022.

\bibitem{wang2024revisiting}
Z.~Wang, C.~Pei, M.~Ma, X.~Wang, Z.~Li, D.~Pei, S.~Rajmohan, D.~Zhang, Q.~Lin,
  H.~Zhang \emph{et~al.}, ``Revisiting vae for unsupervised time series anomaly
  detection: A frequency perspective,'' in \emph{Proceedings of the ACM on Web
  Conference 2024}, 2024, pp. 3096--3105.

\bibitem{fang2023unsupervised}
L.~Fang, K.~Feng, K.~Zhao, A.~Hu, and T.~Li, ``Unsupervised rumor detection
  based on propagation tree vae,'' \emph{IEEE Transactions on Knowledge and
  Data Engineering}, vol.~35, no.~10, pp. 10\,309--10\,323, 2023.

\bibitem{Bian2020Rumor}
T.~Bian, X.~Xiao, T.~Xu, P.~Zhao, W.~Huang, Y.~Rong, and J.~Huang, ``Rumor
  detection on social media with bi-directional graph convolutional networks,''
  in \emph{Proceedings of the AAAI conference on artificial intelligence},
  2020, pp. 549--556.

\bibitem{wei2024modeling}
L.~Wei, D.~Hu, W.~Zhou, X.~Wang, and S.~Hu, ``Modeling the uncertainty of
  information propagation for rumor detection: A neuro-fuzzy approach,''
  \emph{IEEE Transactions on Neural Networks and Learning Systems}, vol.~35,
  no.~2, pp. 2522--2533, 2024.

\bibitem{deng2023markov}
L.~Deng, C.~Wu, D.~Lian, Y.~Wu, and E.~Chen, ``Markov-driven graph
  convolutional networks for social spammer detection,'' \emph{IEEE
  Transactions on Knowledge and Data Engineering}, vol.~35, no.~12, pp.
  12\,310--12\,322, 2023.

\bibitem{Yang2024Topic}
Z.~Yang, Y.~Pang, X.~Li, Q.~Li, S.~Wei, R.~Wang, and Y.~Xiao, ``Topic
  audiolization: A model for rumor detection inspired by lie detection
  technology,'' \emph{Information Processing \& Management}, vol.~61, no.~1, p.
  103563, 2024.

\bibitem{ma2016detecting}
J.~Ma, W.~Gao, P.~Mitra, S.~Kwon, B.~J. Jansen, K.-F. Wong, and M.~Cha,
  ``Detecting rumors from microblogs with recurrent neural networks,'' in
  \emph{Proceedings of the 25th International Joint Conference on Artificial
  Intelligence}, 2016, pp. 3818--3824.

\bibitem{ma2021improving}
J.~Ma, J.~Li, W.~Gao, Y.~Yang, and K.-F. Wong, ``Improving rumor detection by
  promoting information campaigns with transformer-based generative adversarial
  learning,'' \emph{IEEE Transactions on Knowledge and Data Engineering},
  vol.~35, no.~3, pp. 2657--2670, 2023.

\bibitem{Sun2022ddgcn}
M.~Sun, X.~Zhang, J.~Zheng, and G.~Ma, ``Ddgcn: Dual dynamic graph
  convolutional networks for rumor detection on social media,'' in
  \emph{Proceedings of the AAAI Conference on Artificial Intelligence}, 2022,
  pp. 4611--4619.

\bibitem{wang2023cross}
L.~Wang, C.~Zhang, H.~Xu, Y.~Xu, X.~Xu, and S.~Wang, ``Cross-modal contrastive
  learning for multimodal fake news detection,'' in \emph{Proceedings of the
  31st ACM international conference on multimedia}, 2023, pp. 5696--5704.

\bibitem{zhang2024reinforced}
L.~Zhang, X.~Zhang, Z.~Zhou, F.~Huang, and C.~Li, ``Reinforced adaptive
  knowledge learning for multimodal fake news detection,'' in \emph{Proceedings
  of the AAAI Conference on Artificial Intelligence}, vol.~38, no.~15, 2024,
  pp. 16\,777--16\,785.

\bibitem{qu2024temporal}
Z.~Qu, F.~Zhou, X.~Song, R.~Ding, L.~Yuan, and Q.~Wu, ``Temporal enhanced
  multimodal graph neural networks for fake news detection,'' \emph{IEEE
  Transactions on Computational Social Systems}, 2024.

\bibitem{Yang2024model}
Z.~Yang, Y.~Pang, Q.~Li, S.~Wei, R.~Wang, and Y.~Xiao, ``A model for early
  rumor detection base on topic-derived domain compensation and multi-user
  association,'' \emph{Expert Systems with Applications}, vol. 250, p. 123951,
  2024.

\bibitem{qi2023fakesv}
P.~Qi, Y.~Bu, J.~Cao, W.~Ji, R.~Shui, J.~Xiao, D.~Wang, and T.-S. Chua,
  ``Fakesv: A multimodal benchmark with rich social context for fake news
  detection on short video platforms,'' in \emph{Proceedings of the AAAI
  Conference on Artificial Intelligence}, 2023, pp. 14\,444--14\,452.

\end{thebibliography}


\begin{thebibliography}{38}
\bibitem{li2019spam}
A. Li, Z. Qin, R. Liu, Y. Yang, and D. Li, “Spam review detection with graph convolutional networks,” in Proceedings of the 28th ACM international conference on information and knowledge management, 2019, pp. 2703–2711.

\bibitem{zhang2023detecting}
F. Zhang, J. Wu, P. Zhang, R. Ma, and H. Yu, “Detecting collusive spammers with heterogeneous graph attention network,” Information Processing \& Management, vol. 60, no. 3, p. 103282, 2023.

\bibitem{jiang2024learning}
B. Jiang, Z. Zhang, S. Ge, B. Wang, X. Wang, and J. Tang, “Learning graph attentions via replicator dynamics,” IEEE Transactions on Pattern Analysis and Machine Intelligence, vol. 46, no. 12, pp. 7720–7727, 2024.

\bibitem{chen2024gnn}
C. Chen, Y. Wu, Q. Dai, H.-Y. Zhou, M. Xu, S. Yang, X. Han, and Y. Yu, “A survey on graph neural networks and graph transformers in computer vision: A task-oriented perspective,” IEEE Transactions on Pattern Analysis and Machine Intelligence, vol. 46, no. 12, pp. 10297--10318, 2024.

\bibitem{zhang2024predicting}
X. Zhang and W. Gao, “Predicting viral rumors and vulnerable users with graph-based neural multi-task learning for infodemic surveillance,” Information Processing \& Management, vol. 61, no. 1, p. 103520, 2024.

\bibitem{devlin2018bert}
J.~Devlin, ``Bert: Pre-training of deep bidirectional transformers for language
  understanding,'' \emph{arXiv preprint arXiv:1810.04805}, 2018.

\bibitem{radford2019language}
A.~Radford, J.~Wu, R.~Child, D.~Luan, D.~Amodei, I.~Sutskever \emph{et~al.},
  ``Language models are unsupervised multitask learners,'' \emph{OpenAI blog},
  vol.~1, no.~8, p.~9, 2019.

\bibitem{brown2020language}
T.~B. Brown, ``Language models are few-shot learners,'' \emph{arXiv preprint
  arXiv:2005.14165}, 2020.

\bibitem{beltagy2020longformer}
I.~Beltagy, M.~E. Peters, and A.~Cohan, ``Longformer: The long-document
  transformer,'' \emph{arXiv preprint arXiv:2004.05150}, 2020.

\bibitem{achiam2023gpt}
J.~Achiam, S.~Adler, S.~Agarwal, L.~Ahmad, I.~Akkaya, F.~L. Aleman, D.~Almeida,
  J.~Altenschmidt, S.~Altman, S.~Anadkat \emph{et~al.}, ``Gpt-4 technical
  report,'' \emph{arXiv preprint arXiv:2303.08774}, 2023.

\bibitem{wang2024utilizing}
J.~Wang, J.~X. Huang, X.~Tu, J.~Wang, A.~J. Huang, M.~T.~R. Laskar, and
  A.~Bhuiyan, ``Utilizing bert for information retrieval: Survey, applications,
  resources, and challenges,'' \emph{ACM Computing Surveys}, vol.~56, no.~7,
  pp. 1--33, 2024.

\bibitem{liu2021swin}
Z.~Liu, Y.~Lin, Y.~Cao, H.~Hu, Y.~Wei, Z.~Zhang, S.~Lin, and B.~Guo, ``Swin
  transformer: Hierarchical vision transformer using shifted windows,'' in
  \emph{Proceedings of the IEEE/CVF international conference on computer
  vision}, 2021, pp. 10\,012--10\,022.

\bibitem{han2022survey}
K.~Han, Y.~Wang, H.~Chen, X.~Chen, J.~Guo, Z.~Liu, Y.~Tang, A.~Xiao, C.~Xu,
  Y.~Xu \emph{et~al.}, ``A survey on vision transformer,'' \emph{IEEE
  transactions on pattern analysis and machine intelligence}, vol.~45, no.~1,
  pp. 87--110, 2022.

\bibitem{wang2024revisiting}
Z.~Wang, C.~Pei, M.~Ma, X.~Wang, Z.~Li, D.~Pei, S.~Rajmohan, D.~Zhang, Q.~Lin,
  H.~Zhang \emph{et~al.}, ``Revisiting vae for unsupervised time series anomaly
  detection: A frequency perspective,'' in \emph{Proceedings of the ACM on Web
  Conference 2024}, 2024, pp. 3096--3105.

\bibitem{fang2023unsupervised}
L.~Fang, K.~Feng, K.~Zhao, A.~Hu, and T.~Li, ``Unsupervised rumor detection
  based on propagation tree vae,'' \emph{IEEE Transactions on Knowledge and
  Data Engineering}, vol.~35, no.~10, pp. 10\,309--10\,323, 2023.

\bibitem{Bian2020Rumor}
T.~Bian, X.~Xiao, T.~Xu, P.~Zhao, W.~Huang, Y.~Rong, and J.~Huang, ``Rumor
  detection on social media with bi-directional graph convolutional networks,''
  in \emph{Proceedings of the AAAI conference on artificial intelligence},
  2020, pp. 549--556.

\bibitem{wei2024modeling}
L.~Wei, D.~Hu, W.~Zhou, X.~Wang, and S.~Hu, ``Modeling the uncertainty of
  information propagation for rumor detection: A neuro-fuzzy approach,''
  \emph{IEEE Transactions on Neural Networks and Learning Systems}, vol.~35,
  no.~2, pp. 2522--2533, 2024.

\bibitem{deng2023markov}
L.~Deng, C.~Wu, D.~Lian, Y.~Wu, and E.~Chen, ``Markov-driven graph
  convolutional networks for social spammer detection,'' \emph{IEEE
  Transactions on Knowledge and Data Engineering}, vol.~35, no.~12, pp.
  12\,310--12\,322, 2023.

\bibitem{Yang2024Topic}
Z.~Yang, Y.~Pang, X.~Li, Q.~Li, S.~Wei, R.~Wang, and Y.~Xiao, ``Topic
  audiolization: A model for rumor detection inspired by lie detection
  technology,'' \emph{Information Processing \& Management}, vol.~61, no.~1, p.
  103563, 2024.

\bibitem{ma2016detecting}
J.~Ma, W.~Gao, P.~Mitra, S.~Kwon, B.~J. Jansen, K.-F. Wong, and M.~Cha,
  ``Detecting rumors from microblogs with recurrent neural networks,'' in
  \emph{Proceedings of the 25th International Joint Conference on Artificial
  Intelligence}, 2016, pp. 3818--3824.

\bibitem{ma2021improving}
J.~Ma, J.~Li, W.~Gao, Y.~Yang, and K.-F. Wong, ``Improving rumor detection by
  promoting information campaigns with transformer-based generative adversarial
  learning,'' \emph{IEEE Transactions on Knowledge and Data Engineering},
  vol.~35, no.~3, pp. 2657--2670, 2023.

\bibitem{Sun2022ddgcn}
M.~Sun, X.~Zhang, J.~Zheng, and G.~Ma, ``Ddgcn: Dual dynamic graph
  convolutional networks for rumor detection on social media,'' in
  \emph{Proceedings of the AAAI Conference on Artificial Intelligence}, 2022,
  pp. 4611--4619.

\bibitem{wang2023cross}
L.~Wang, C.~Zhang, H.~Xu, Y.~Xu, X.~Xu, and S.~Wang, ``Cross-modal contrastive
  learning for multimodal fake news detection,'' in \emph{Proceedings of the
  31st ACM international conference on multimedia}, 2023, pp. 5696--5704.

\bibitem{zhang2024reinforced}
L.~Zhang, X.~Zhang, Z.~Zhou, F.~Huang, and C.~Li, ``Reinforced adaptive
  knowledge learning for multimodal fake news detection,'' in \emph{Proceedings
  of the AAAI Conference on Artificial Intelligence}, vol.~38, no.~15, 2024,
  pp. 16\,777--16\,785.

\bibitem{qu2024temporal}
Z.~Qu, F.~Zhou, X.~Song, R.~Ding, L.~Yuan, and Q.~Wu, ``Temporal enhanced
  multimodal graph neural networks for fake news detection,'' \emph{IEEE
  Transactions on Computational Social Systems}, 2024.

\bibitem{Yang2024model}
Z.~Yang, Y.~Pang, Q.~Li, S.~Wei, R.~Wang, and Y.~Xiao, ``A model for early
  rumor detection base on topic-derived domain compensation and multi-user
  association,'' \emph{Expert Systems with Applications}, vol. 250, p. 123951,
  2024.

\bibitem{zhang2023rumor}
Q. Zhang, Y. Yang, C. Shi, A. Lao, L. Hu, S. Wang, and U. Naseem, “Rumor detection with hierarchical representation on bipartite ad hocevent trees,” IEEE Transactions on Neural Networks and Learning Systems, vol. 35, no. 10, pp. 14x112–-14124, 2024.

\bibitem{babu2023efficient}
R. Babu, J. Kannappan, B. V. Krishna, and K. Vijay, “An efficient spam detector model for accurate categorization of spam tweets using quantum chaotic optimization-based stacked recurrent network,” Nonlinear Dynamics, vol. 111, no. 19, pp. 18523–-18540, 2023.

\bibitem{GRU}
K. Yu, X. Zhu, Z. Guo, A. Tolba, J. J. P. C. Rodrigues, and V. C. Leung, “A cross-field deep learning-based fuzzy spamming detection approach via collaboration of behavior modeling and sentiment analysis,” IEEE Transactions on Fuzzy Systems, vol. 32, no. 12, pp. 7168–7182, 2024.

\bibitem{rao2023hybrid}
S. Rao, A. K. Verma, and T. Bhatia, “Hybrid ensemble framework with self-attention mechanism for social spam detection on imbalanced data,” Expert Systems with Applications, vol. 217, p. 119594, 2023.

\bibitem{gao2019graph}
H. Gao and S. Ji, “Graph u-nets,” in international conference on machine learning. PMLR, 2019, pp. 2083–-2092.

\bibitem{generale2022scaling}
A. Generale, T. Blume, and M. Cochez, “Scaling r-gcn training with graph summarization,” in Companion Proceedings of the Web Conference 2022, 2022, pp. 1073–-1082.

\bibitem{he2022convolutional}
M. He, Z. Wei, and J.-R. Wen, “Convolutional neural networks on graphs with chebyshev approximation, revisited,” Advances in neural information processing systems, vol. 35, pp. 7264–-7276, 2022.

\bibitem{GraphTrans}
Y. Cai, H. Wang, H. Cao, W. Wang, L. Zhang, and X. Chen, “Detecting spam movie review under coordinated attack with multi-view explicit and implicit relations semantics fusion,” IEEE Transactions on Information Forensics and Security, vol. 19, pp. 7588–-7603, 2024.

\bibitem{zhang2022detecting}
F. Zhang, S. Yuan, P. Zhang, J. Chao, and H. Yu, “Detecting review spammer groups based on generative adversarial networks,” Information Sciences, vol. 606, pp. 819–836, 2022.

\bibitem{yang2022chinese}
A. Yang, J. Pan, J. Lin, R. Men, Y. Zhang, J. Zhou, and C. Zhou, “Chinese clip: Contrastive vision-language pretraining in chinese,” arXiv preprint arXiv:2211.01335, 2022.

\bibitem{jia2021scaling}
C. Jia, Y. Yang, Y. Xia, Y.-T. Chen, Z. Parekh, H. Pham, Q. Le, Y.-H. Sung, Z. Li, and T. Duerig, “Scaling up visual and vision-language representation learning with noisy text supervision,” in International conference on machine learning. PMLR, 2021, pp. 4904–4916.

\bibitem{li2023blip}
J. Li, D. Li, S. Savarese, and S. Hoi, “Blip-2: Bootstrapping languageimage pre-training with frozen image encoders and large language models,” in International conference on machine learning. PMLR, 2023, pp. 19 730–19 742.
\end{thebibliography}
\vspace{-15mm}
\begin{IEEEbiography}[{\includegraphics[width=1in,height=1.2in,clip,keepaspectratio]{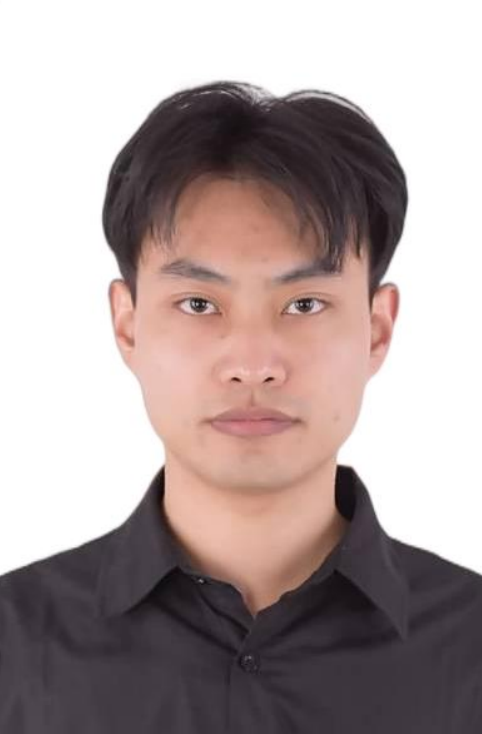}}]{Zhou Yang}
	received the B.S degree in electronic Science and Technology from Zhengzhou University of Science and Technology, China, in 2021. He is currently working toward the M.S degree in information and communication engineering with Chongqing University of Posts and Telecommunications, China. His research interests include rumor detection, spammer detection, deep learning, and its application.
\end{IEEEbiography}
\vspace{-15mm}
\begin{IEEEbiography}[{\includegraphics[width=1in,height=1.2in,clip,keepaspectratio]{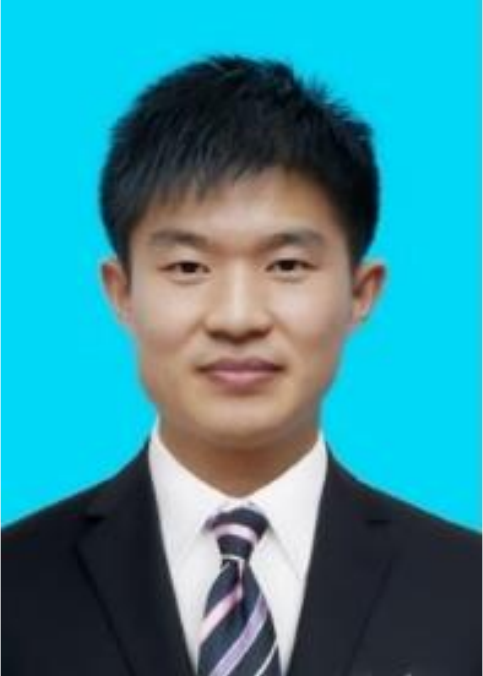}}]{Yucai Pang}
	received the Ph.D. degree from the Harbin Engineering University, Harbin, China. He has been working as an Associate Professor with the School of Communication and Information Engineering, Chongqing University of Posts and Telecommunications since 2017. His research interests include big data prediction, array signal processing, and social rumor detection.
\end{IEEEbiography}
\vspace{-15mm}
\begin{IEEEbiography}[{\includegraphics[width=1in,height=1.2in,clip,keepaspectratio]{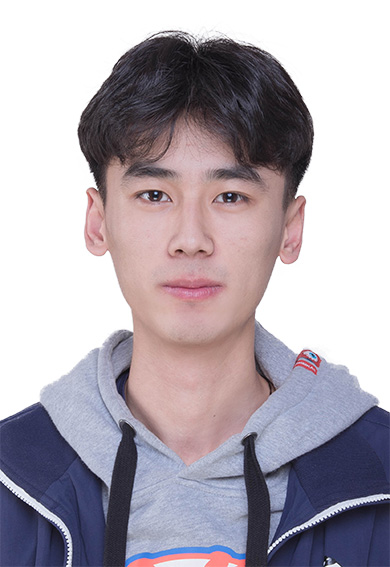}}]{Hongbo Yin}
	is currently working toward his doctoral degree in the School of Information and Communication Engineering, University of Electronic Science and Technology of China (UESTC), Chengdu, China. He received the M.Sc. degree in Chongqing University of Posts and Telecommunications, Chongqing, China, in 2024. His main research interests are social networks, spammer detection, edge computing networks.
\end{IEEEbiography}
\vspace{-15mm}
\begin{IEEEbiography}[{\includegraphics[width=1in,height=1.2in,clip,keepaspectratio]{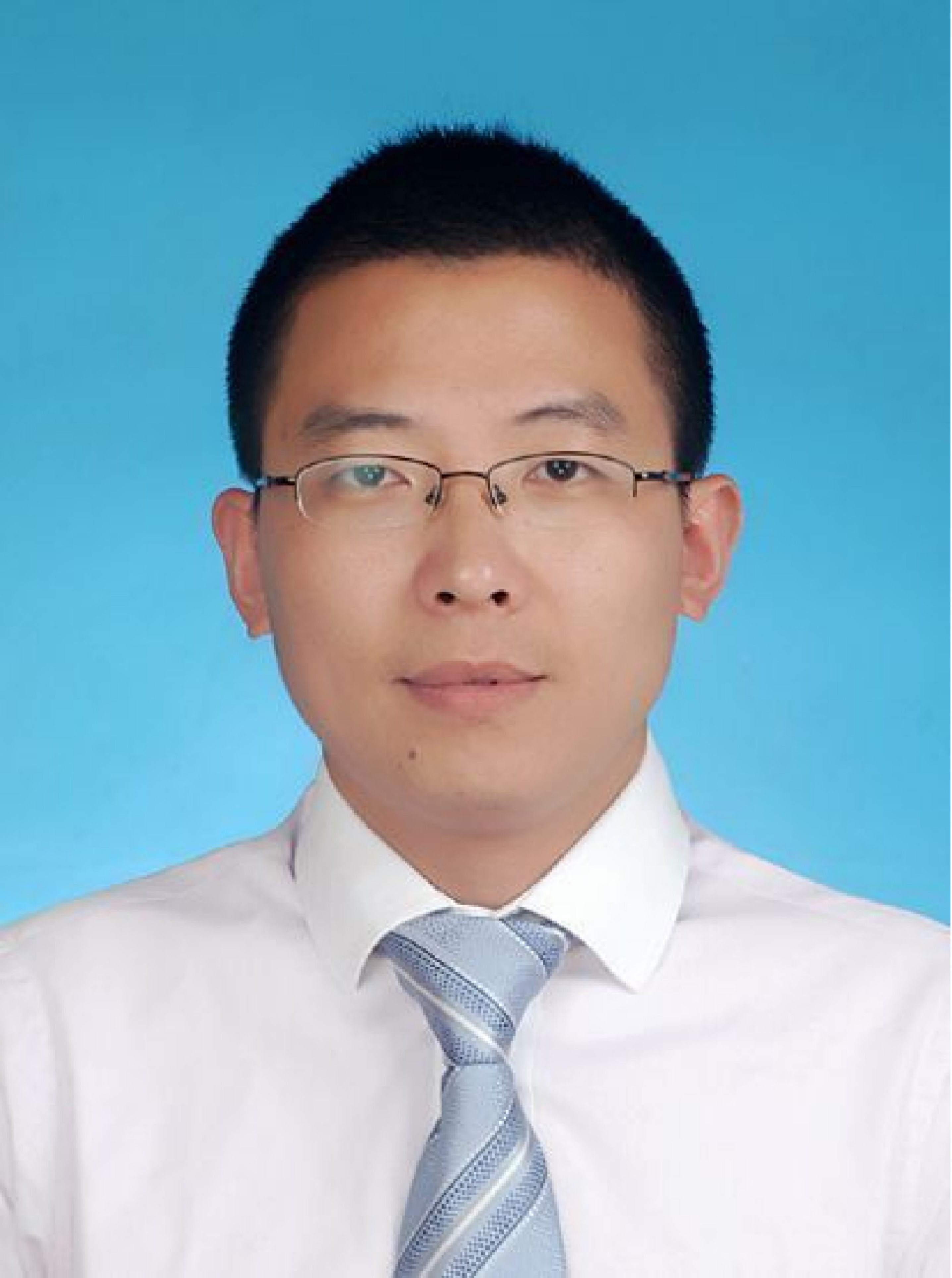}}]{Yunpeng Xiao}
	received the Ph.D. degree in computer science from Beijing University of Posts and Telecommunications, Beijing, China, in 2013. He is a professor and vice dean of the Institute of Electronic Information and Network Engineering, Chongqing University of Posts and Telecommunications, Chongqing, China. He was a visiting scholar of Georgia Institute of Technology from 2018 to 2019. His research interests include social networks, e-commerce and intelligent system.
\end{IEEEbiography}

\end{document}